\documentclass[10pt,twocolumn,letterpaper]{article}

\usepackage[pagenumbers]{iccv} 
\usepackage{multirow}
\usepackage{graphicx}
\usepackage{array}
\usepackage{ragged2e} 
\definecolor{iccvblue}{rgb}{0.21,0.49,0.74}
\usepackage[pagebackref,breaklinks,colorlinks,allcolors=iccvblue]{hyperref}
\usepackage{float}
\def\paperID{10070} 
\def\confName{ICCV}
\def\confYear{2025}

\begin{document}

\title{Mitigating Pretraining-Induced Attention Asymmetry in 2D+ Electron Microscopy Image Segmentation
}

\author{Zsófia Molnár\\
{\tt\small molnar.zsofia@itk.ppke.hu}
\and
Gergely Szabó\\
{\tt\small szabo.gergely@itk.ppke.hu}
\and
András Horváth\\
{\tt\small horvath.andras@itk.ppke.hu}
\and
ITK, PPCU\\
Budapest, Pr\'ater st. 50/A, 1083, Hungary
}

\maketitle


\begin{abstract}
Vision models pretrained on large-scale RGB natural image datasets are widely reused for electron microscopy image segmentation. In electron microscopy, volumetric data are acquired as serial sections and processed as stacks of adjacent grayscale slices, where neighboring slices provide symmetric contextual information for identifying features on the central slice. The common strategy maps such stacks to pseudo-RGB inputs to enable transfer learning from pretrained models. However, this mapping imposes channel-specific semantics inherited from natural images, even though electron microscopy slices are homogeneous in the modality and symmetric in their predictive roles. As a result, pretrained models may encode inductive biases that are misaligned with the inherent symmetry of volumetric electron microscopy data.

In this work, it is demonstrated that RGB-pretrained models systematically assign unequal importance to individual input slices when applied to stacked electron microscopy data, despite the absence of any intrinsic channel ordering. Using saliency-based attribution analysis across multiple architectures, a consistent channel-level asymmetry was observed that persists after fine-tuning and affects model interpretability, even when segmentation performance is unchanged. To address this issue, a targeted modification of pretraining weights based on uniform channel initialization was proposed, which restores symmetric feature attribution while preserving the benefits of pretraining. Experiments on the SNEMI, Lucchi and GF-PA66 datasets confirm a substantial reduction in attribution bias without compromising or even improving segmentation accuracy.

\end{abstract}


\section{Introduction} \label{sec:intro}

Transfer learning with pretrained neural networks is a standard approach in image analysis, enabling strong performance with limited labeled data by reusing representations learned from large-scale datasets \cite{yosinski2014transferable, kornblith2019better}. Although effective in many settings, pretrained models may encode assumptions that conflict with the structural properties of specialized imaging domains \cite{renggli2022model, zhuang2020comprehensive}. This mismatch is particularly relevant in electron microscopy, where volumetric data is acquired as serial sections and is commonly processed as stacks of adjacent grayscale slices. To utalize RGB-pretrained models, neighboring slices are frequently mapped to pseudo-RGB inputs \cite{vu2020evaluation, zhang2022bridging, avesta2023comparing}, with the central slice serving as the prediction target. In such representations, however, adjacent slices correspond to symmetric spatial cross-sections of the same three-dimensional structure and should have no inherent channel-specific meaning. Consequently, a faithful model representation should treat these input channels equivalently.

In contrast, vision models pretrained on natural RGB images, including convolutional and transformer-based architectures, encode channel-specific feature preferences shaped by color statistics that are unrelated to electron microscopy data \cite{rafegas2018color, de2022emergent, bahador2025vision}. When such pretrained weights are applied to stacks of grayscale electron microscopy slices, they induce systematic asymmetries in channel utilization and feature attribution. Although these effects may not substantially impact segmentation accuracy, they undermine interpretability and violate the intrinsic structural symmetry of volumetric electron microscopy data. This issue is particularly consequential in electron microscopy analysis, where segmentation results support downstream quantitative studies of cellular morphology and ultrastructure. Asymmetric feature attribution across slices can propagate into biased or misleading biological interpretations, even when conventional performance metrics remain unchanged. Moreover, the growing emphasis on interpretability in biomedical machine learning underscores the need to better understand how pretrained representations shape model behavior in this domain \cite{dwivedi2023explainable, amann2020explainability}.

\renewcommand{\arraystretch}{1.3}

\begin{table*}[h!]
\centering
\resizebox{0.85\textwidth}{!}{
\begin{tabular}{
>{\raggedright\arraybackslash}p{0.25\textwidth}
p{0.62\textwidth}
}
\hline
\textbf{Problem or Issue} &
Vision models pretrained on RGB natural images are widely applied to stacked grayscale electron microscopy data, but this practice introduces channel-specific biases that conflict with the symmetric structure of serial electron microscopy slices. \\
\hline
\textbf{What is Already Known} &
Transfer learning improves performance in electron microscopy segmentation, yet pretrained models often require adaptation to better align with domain-specific data properties and interpretability requirements. \\
\hline
\textbf{What this Paper Adds} &
Demonstrates systematic channel-level attribution asymmetry induced by RGB-pretrained weights in stacked electron microscopy images, and proposes a simple mitigation strategy that restores symmetric channel importance without performance loss. \\
\hline
\textbf{Who would benefit from the new knowledge in this paper} &
Researchers and practitioners working on electron microscopy segmentation, and developers of interpretable deep learning methods for stacked grayscale imaging data. \\
\hline
\end{tabular}
}
\caption{Statement of Significance}
\label{tab:stat_sign}
\end{table*}

The objective of this study is to systematically characterize channel-level attribution asymmetries induced by RGB-pretrained weights in stacked electron microscopy data. Using saliency-based analysis on established electron microscopy benchmarks, including SNEMI \cite{arganda2013snemi3d} and Lucchi \cite{lucchi2011supervoxel}, a consistent asymmetric channel attention was found across architectures. Nonetheless, a targeted modification of pretrained weights was introduced that restores symmetric channel attribution while preserving predictive performance. The significance of the study is summarized in \Cref{tab:stat_sign}.

\section{Background} \label{sec:background}

\subsection{Pretrained Models and Transfer Learning} \label{sec:pretrain_transfer}

Pretraining on large-scale natural image datasets such as ImageNet has become a cornerstone of modern deep learning, enabling efficient transfer learning by reducing annotation demands and computational cost \cite{ImageNet, othman2022automatic}.  This is particularly true in applications where data is limited, as is often the case in biomedical fields. With the emergence of foundation models in vision, this issue becomes even more pronounced \cite{lee2024foundation}, as these models are pretrained on larger, more generic datasets, predominantly consisting of RGB images. However, pretrained models inevitably inherit inductive biases that reflect the statistical structure of their source domain, which may not align with the target data distribution.

In natural images, RGB channels encode perceptual color information and are treated asymmetrically, both by network architectures and by statistics of human vision. In particular, the green channel contributes disproportionately to luminance perception \cite{wald1964receptors, fairchild2013color}, a bias that is implicitly captured during ImageNet pretraining \cite{deng2009imagenet}. When such models are transferred to domains where the input channels represent structural or spatial context rather than color—such as stacked grayscale electron microscopy slices—these learned channel hierarchies persist despite being semantically inappropriate. This domain-level mismatch provides a mechanistic explanation for the asymmetric channel attribution observed in pretrained models applied to volumetric electron microscopy data, motivating the need for interventions that reconcile pretrained representations with the symmetry of the target domain.

\subsection{Challenges and Benefits of 2D+ Representations in Imaging}

Electron microscopy datasets pose significant challenges due to their high resolution, structural complexity, and limited availability of annotations. These constraints have motivated the widespread adoption of transfer learning, despite the potential mismatches between pretrained models and electron microscopy data characteristics. 

The common strategy for incorporating volumetric context into 2D architectures is the use of 2D+ representations --- often also referred to as 2.5D --- where adjacent slices are assigned to separate input channels (e.g., previous, current, and next slice) \cite{roth2014new, xia2018bridging, xing20192}. This approach enables models to capture a degree of spatial continuity while avoiding the substantial memory and computational costs associated with full 3D convolution. Compared to pure 2D processing, 2D+ representations preserve inter-slice context, which is critical for accurate segmentation of complex structures. At the same time, they remain compatible with existing pretrained 2D models. However, this compatibility comes at the cost of repurposing color channels to encode spatial information, exposing the model to channel-specific biases inherited from RGB pretraining \cite{rafegas2018color, de2022emergent, bahador2025vision}, or even other unique characteristics of macroscopic imaging such as the object positioning bias \cite{szabo2022mitigating, kayhan2020translation}.

Although 2D+ strategies offer a practical compromise between spatial expressiveness and computational efficiency, their interaction with RGB-pretrained weights warrants careful examination, particularly in domains such as electron microscopy where macroscopic homogeneity and thus positional importance symmetry is a fundamental property of the data \cite{de2021deciphering, peddie2022volume, hanslovsky2017image}.

\subsection{Saliency Asymmetry and Attention Mechanisms} \label{sec:saliencymethods}

In the context of stacked grayscale inputs, saliency distributions are expected to reflect the structural symmetry of the representation. Although the central slice typically carries greater relevance due to its direct association with the prediction target, the surrounding slices serve symmetrically equivalent contextual roles. Therefore, systematic differences in their attributed importance indicate asymmetry in how the model processes input channels. Throughout this work, the name \textit{saliency asymmetry} is introduced to refer to such persistent, channel-dependent deviations in feature attribution.

\begin{figure}[h!]
\centering
\resizebox{0.95\columnwidth}{!}{
    \centering
    \raisebox{-1.75cm}{ 
        \begin{subfigure}{0.49\textwidth}
            \centering
            \includegraphics[width=0.835\textwidth]{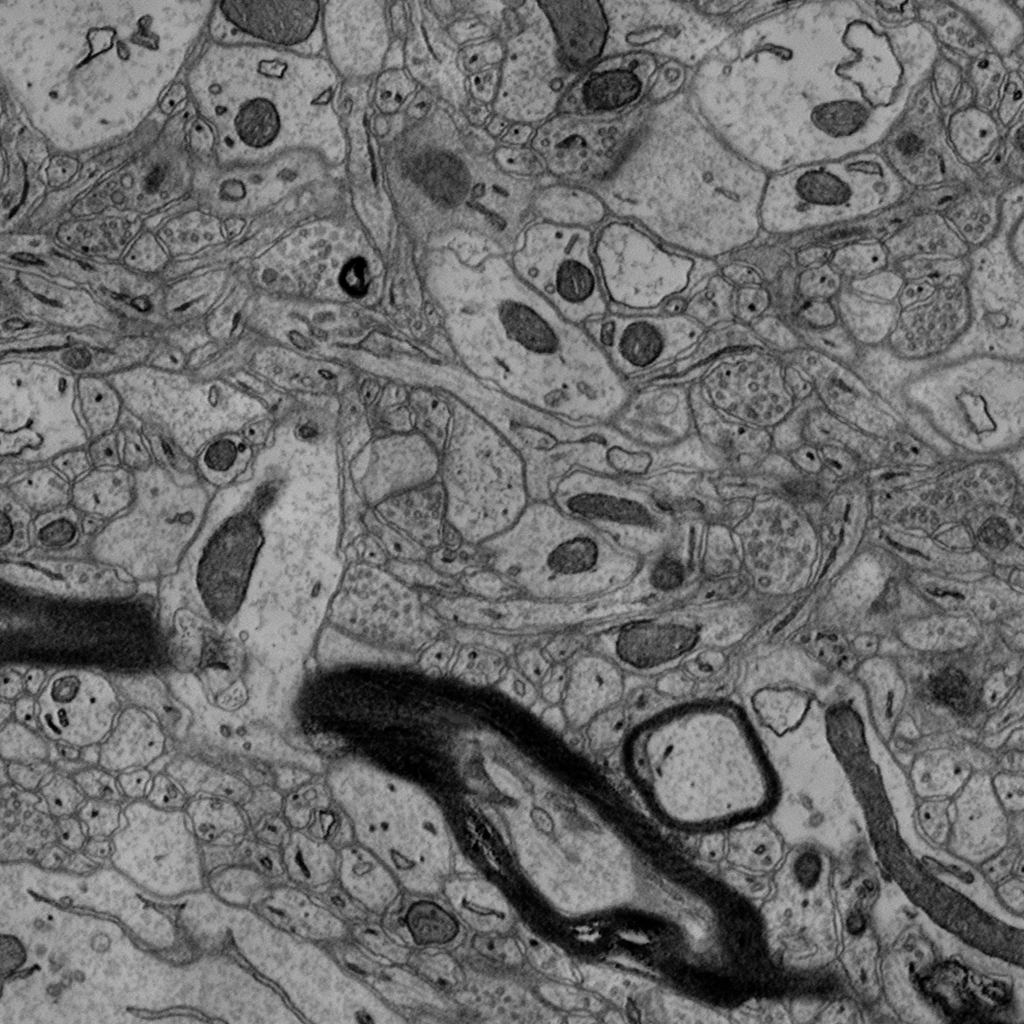}
            \caption{Input Image}
        \end{subfigure}
    }
    \hfill
    \begin{subfigure}{0.58\textwidth}
        \centering
        \begin{minipage}{\textwidth}
            \centering
            \begin{subfigure}{0.32\textwidth}
                \centering
                \includegraphics[width=\textwidth]{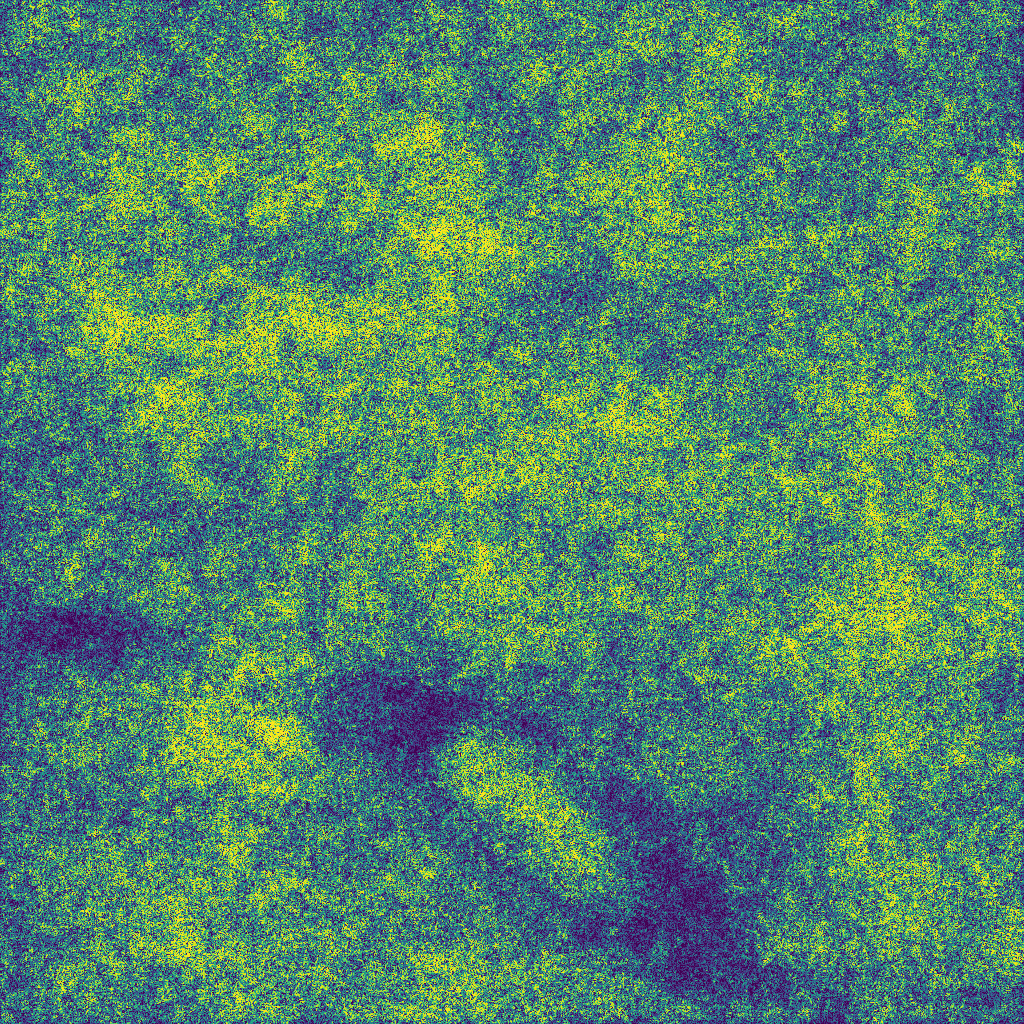}
                \caption{Foreground}
            \end{subfigure}
            \hfill
            \begin{subfigure}{0.32\textwidth}
                \centering
                \includegraphics[width=\textwidth]
                {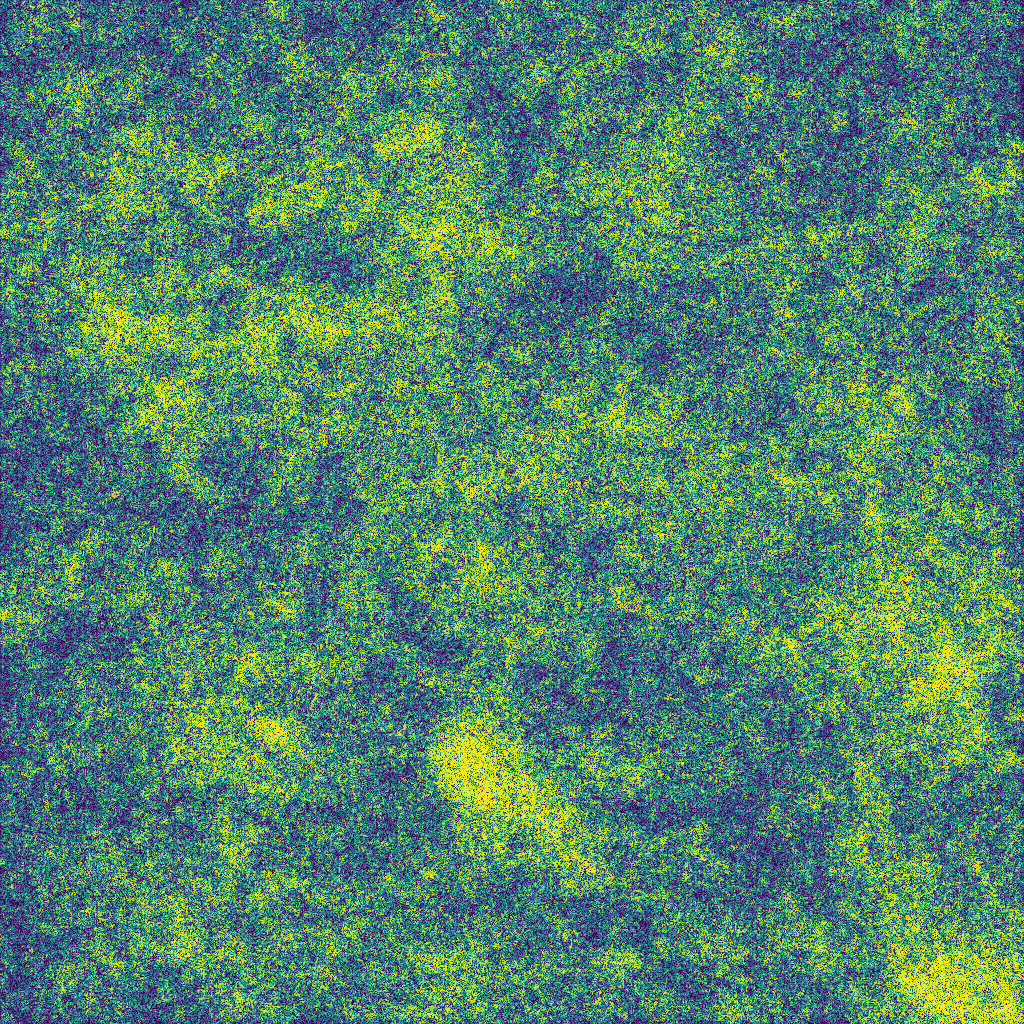}
                \caption{Full Output}
            \end{subfigure}
            \hfill
            \begin{subfigure}{0.32\textwidth}
                \centering
                \includegraphics[width=\textwidth]
                {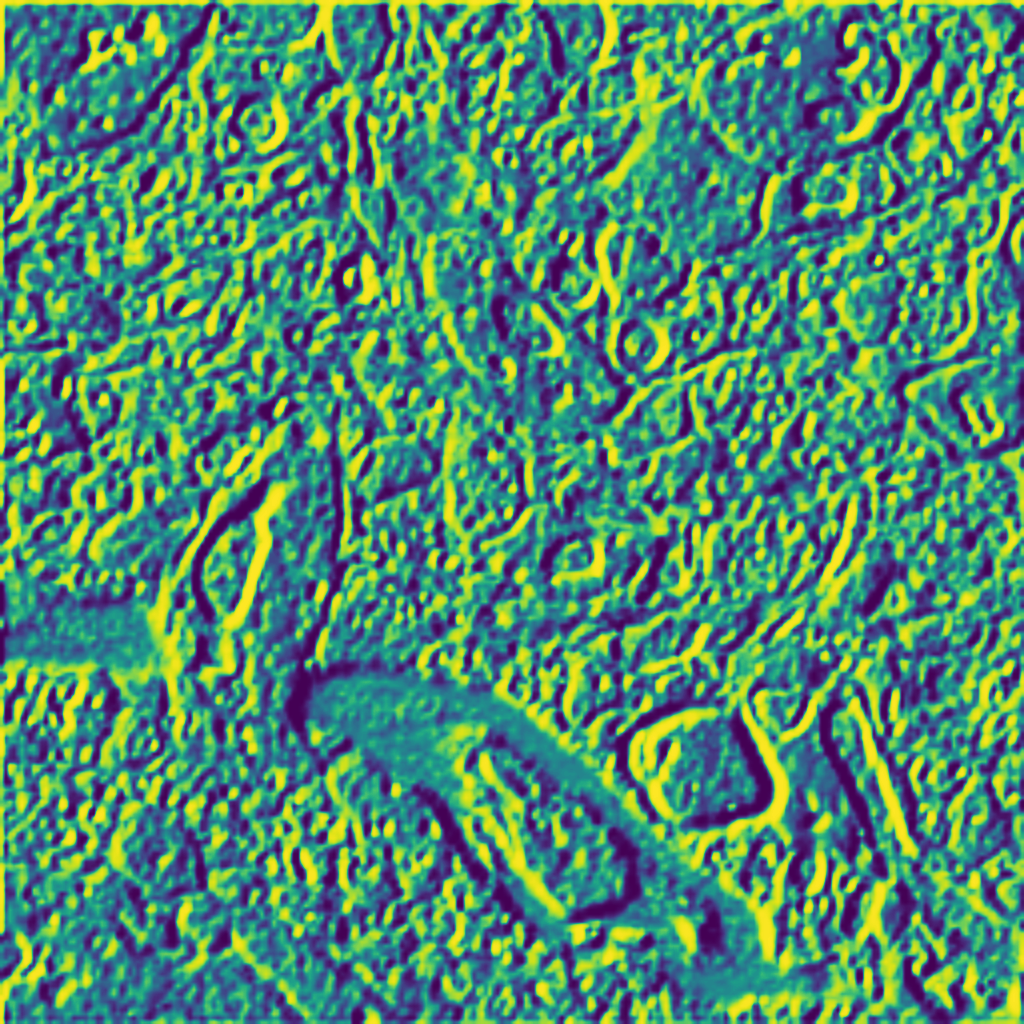}
                \caption{GradCam++}
            \end{subfigure}
        \end{minipage}
        
        \vspace{0.2cm}
        
        \begin{minipage}{\textwidth}
            \centering
            \begin{subfigure}{0.32\textwidth}
                \centering
                \includegraphics[width=\textwidth]
                {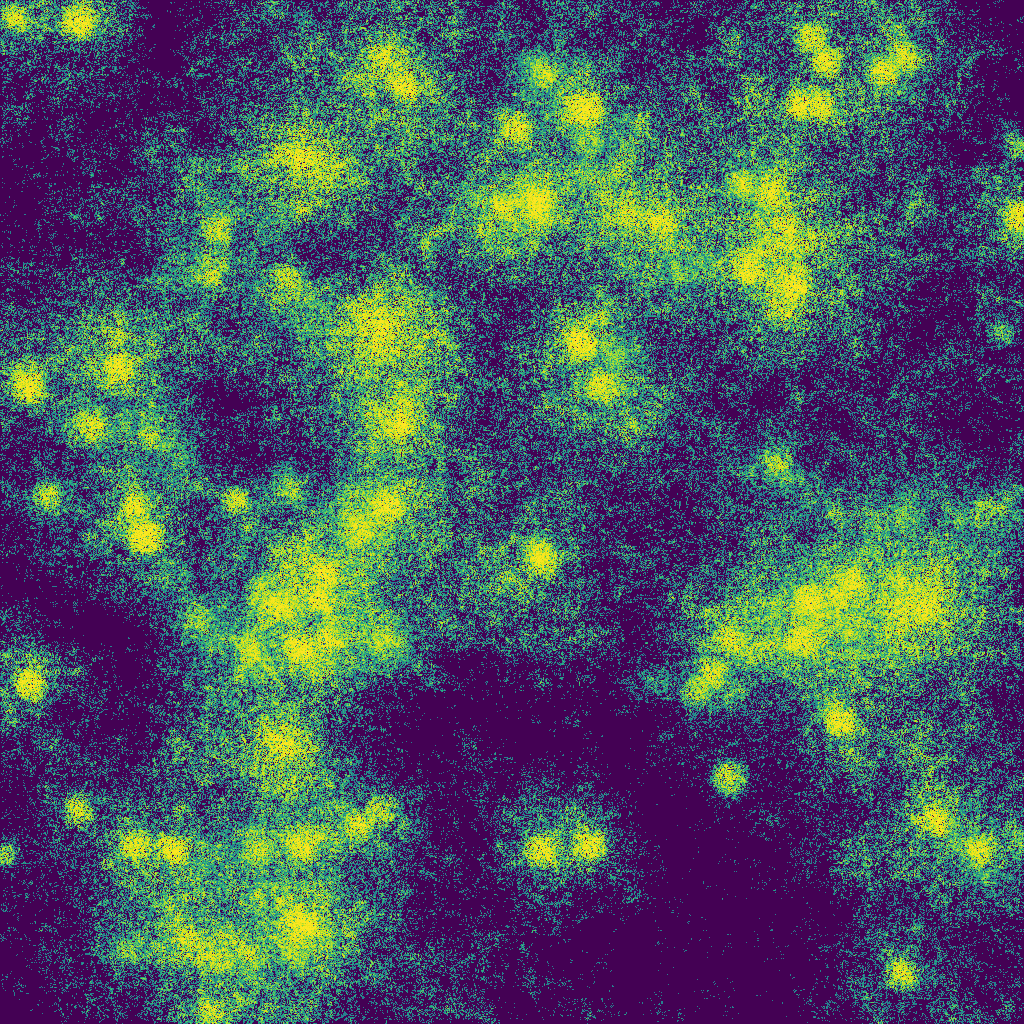}
                \caption{Foreground100}
            \end{subfigure}
            \hfill
            \begin{subfigure}{0.32\textwidth}
                \centering
                \includegraphics[width=\textwidth]
                {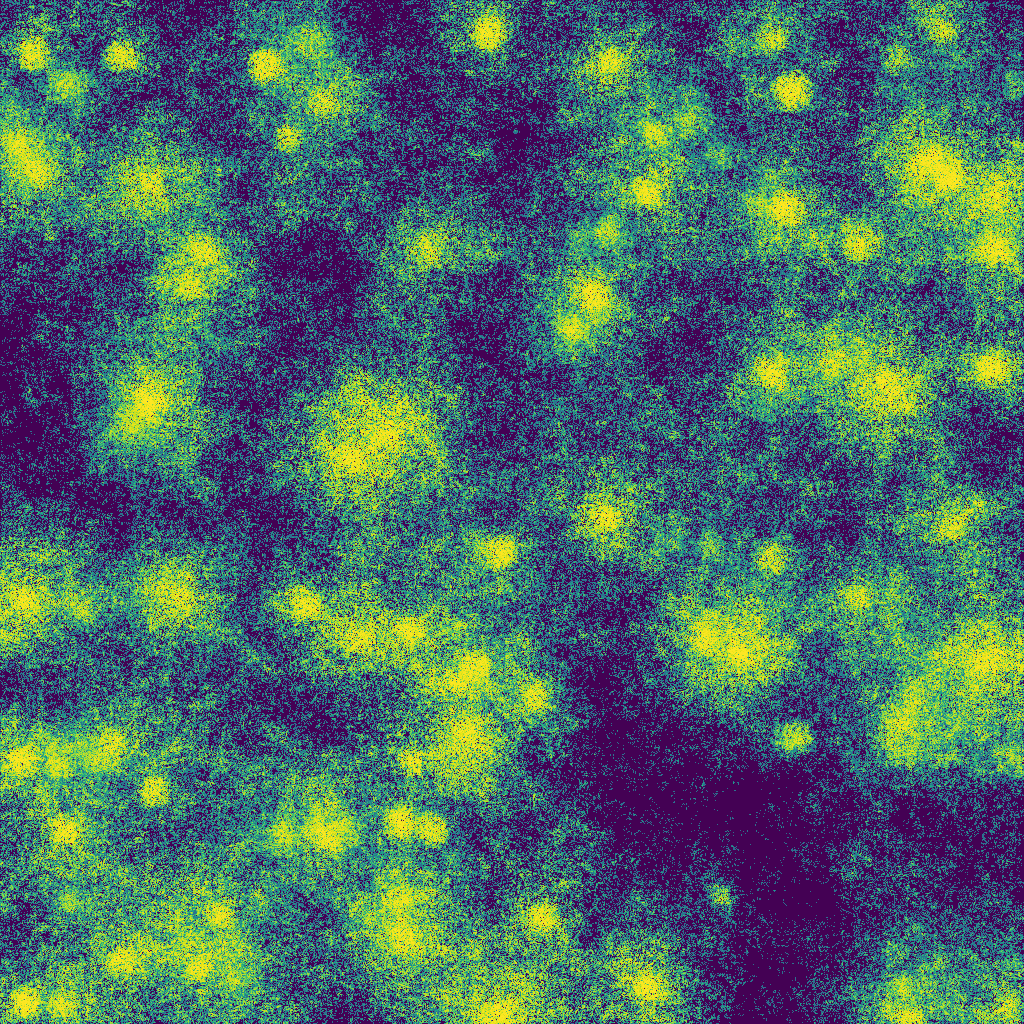}
                \caption{Full Output100}
            \end{subfigure}
            \hfill
            \begin{subfigure}{0.32\textwidth}
                \centering
                \includegraphics[width=\textwidth]{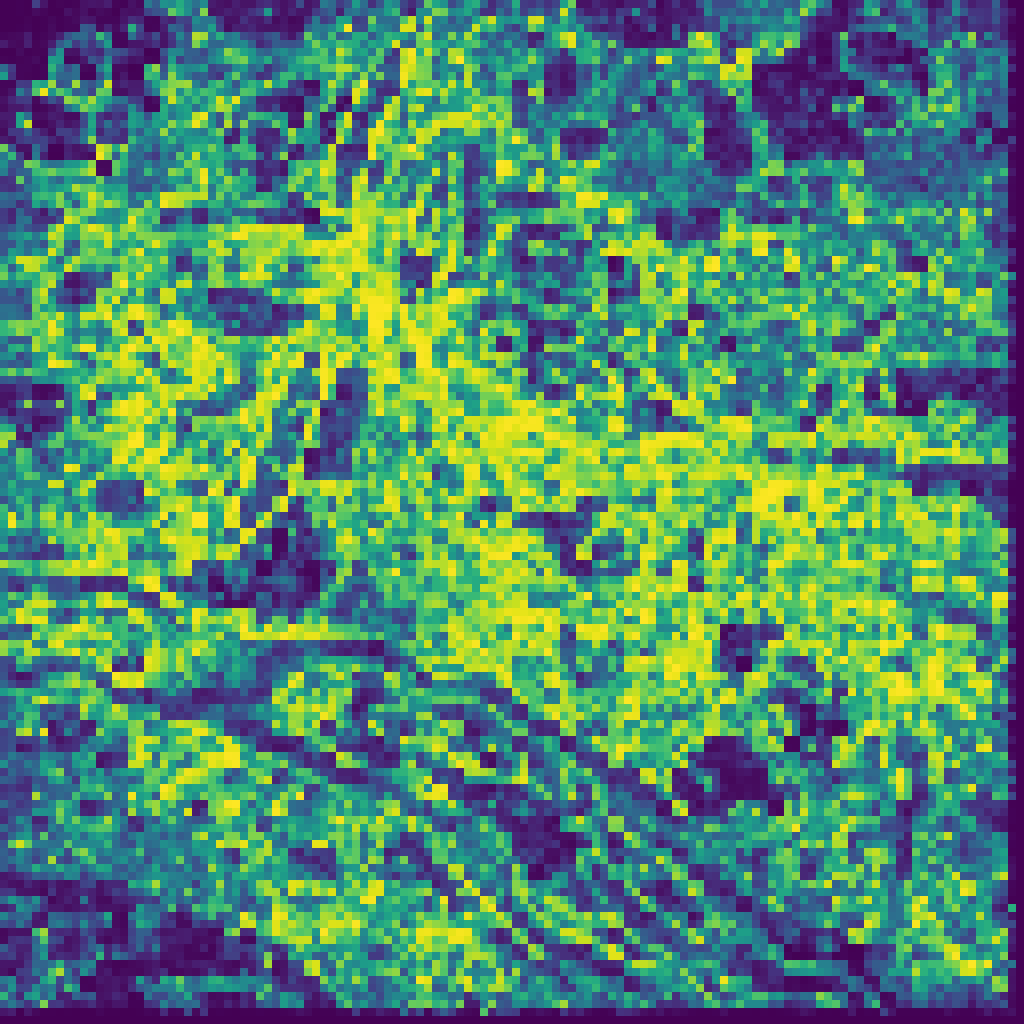}
                \caption{Occlusion}
            \end{subfigure}
        \end{minipage}
    \end{subfigure}
}
\caption{Visualization of a randomly selected SNEMI image alongside various saliency map types used in this study. The saliency map visualizations underwent histogram equalization to ensure a comparable scale across different methods.}
\label{fig:saliency_types}
\end{figure}

For dense prediction tasks, saliency estimation requires aggregation over multiple output locations, as relevance is distributed across spatially extended predictions rather than a single scalar output. In this study, saliency maps were primarily computed using gradient backpropagation with respect to the summed prediction outputs, providing a global estimate of input relevance. To assess whether the observed channel-level effects depended on a specific attribution formulation, several alternative strategies were also evaluated, including foreground-weighted aggregation, randomly sampled output locations, occlusion-based saliency, and channel-occluded variants of GradCAM and GradCAM++ \cite{selvaraju2017grad, chattopadhay2018grad}. Although these methods differ substantially in their assumptions, spatial resolution, and interpretive scope, they capture consistent patterns in the relative distribution of attention across input channels \cite{mokuwe2020black, zeiler2014visualizing}. The convergence of channel-level behavior across attribution methods with distinct computational properties suggests that the observed asymmetry reflects an intrinsic characteristic of the trained models rather than an artifact of a particular saliency technique. Representative examples are shown in \Cref{fig:saliency_types}.

\vspace{-10pt}

\section{Datasets} \label{sec:datasets}

\begin{figure}[h!]
\centering
\resizebox{0.95\columnwidth}{!}{
    \centering
    \begin{subfigure}[b]{.3\columnwidth}
        \centering
        \includegraphics[width=\linewidth]{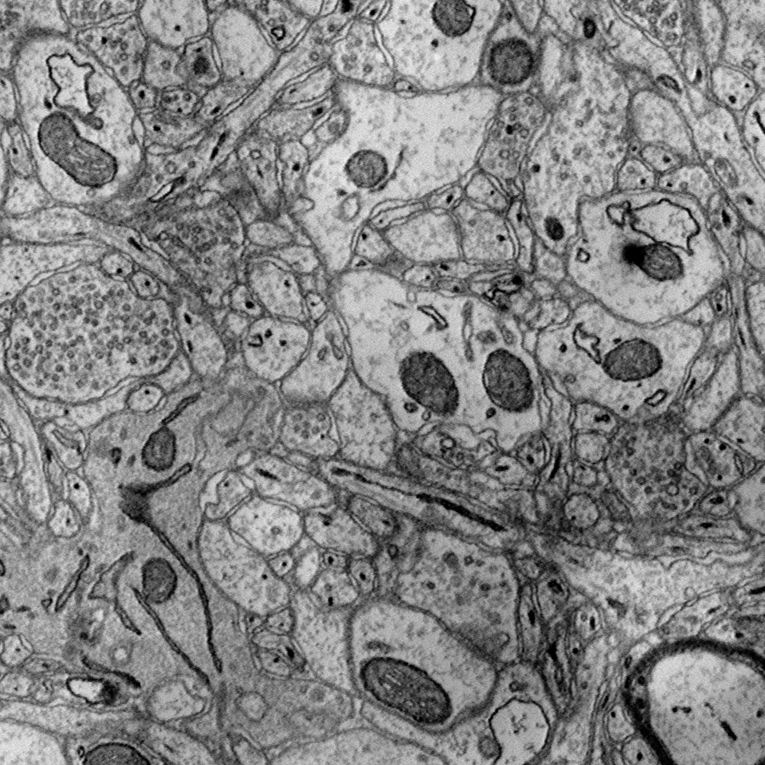}
        \caption{SNEMI raw}
    \end{subfigure}
    \begin{subfigure}[b]{.3\columnwidth}
        \centering
        \includegraphics[width=\linewidth]{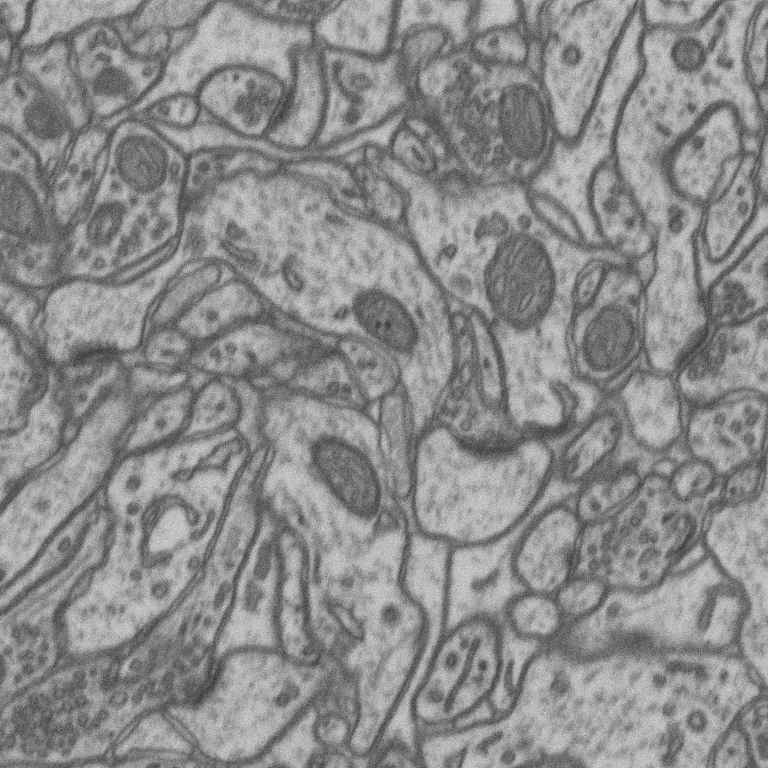}
        \caption{Lucchi raw}
    \end{subfigure}
    \begin{subfigure}[b]{.3\columnwidth}
        \centering
        \includegraphics[width=\linewidth]{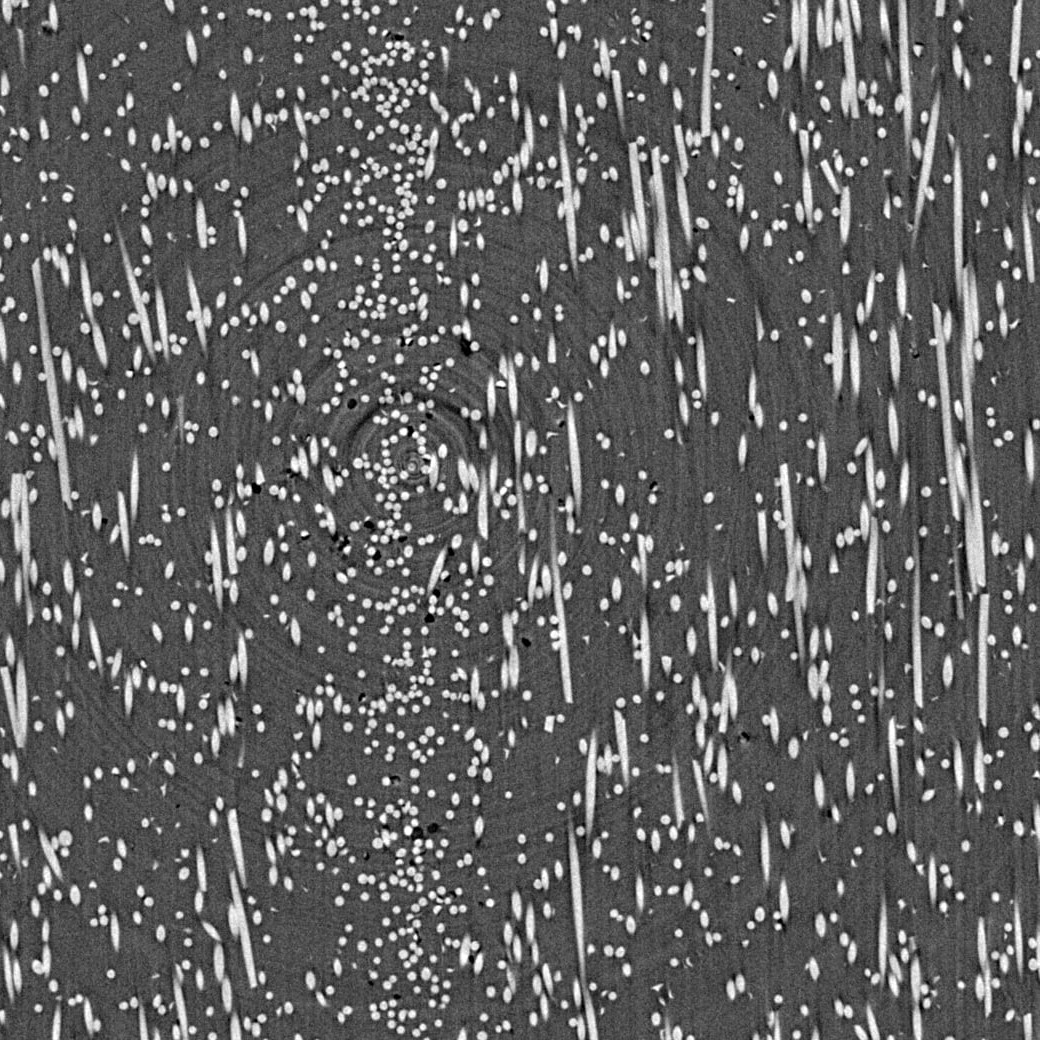}
        \caption{GF-PA66 raw}
    \end{subfigure}}
    
    \vspace{1mm} 
    
   \centering
\resizebox{0.95\columnwidth}{!}{ 
    \begin{subfigure}[b]{.3\columnwidth}
        \centering
        \includegraphics[width=\linewidth]{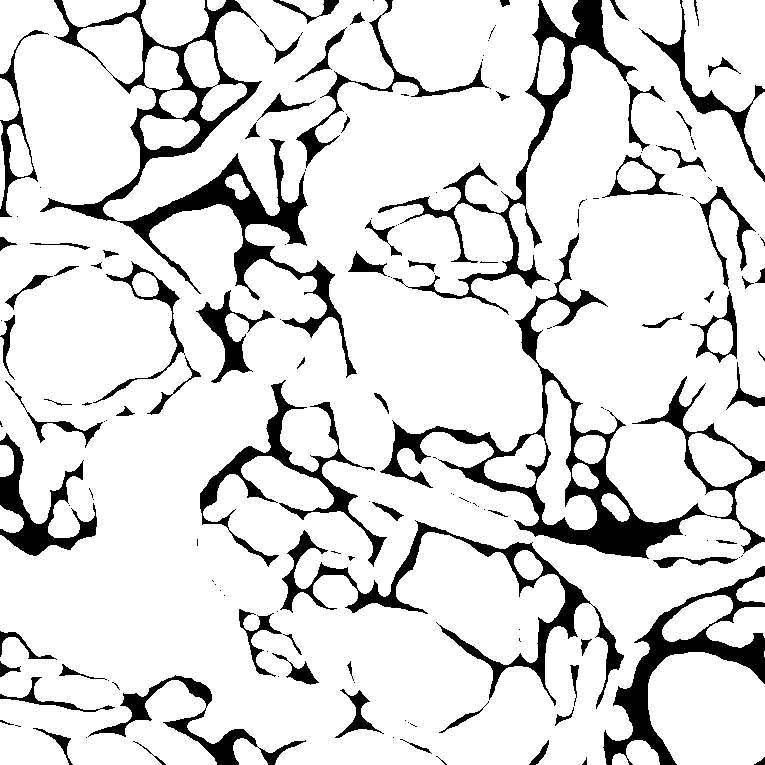}
        \caption{SNEMI mask}
    \end{subfigure}
    \begin{subfigure}[b]{.3\columnwidth}
        \centering
        \includegraphics[width=\linewidth]{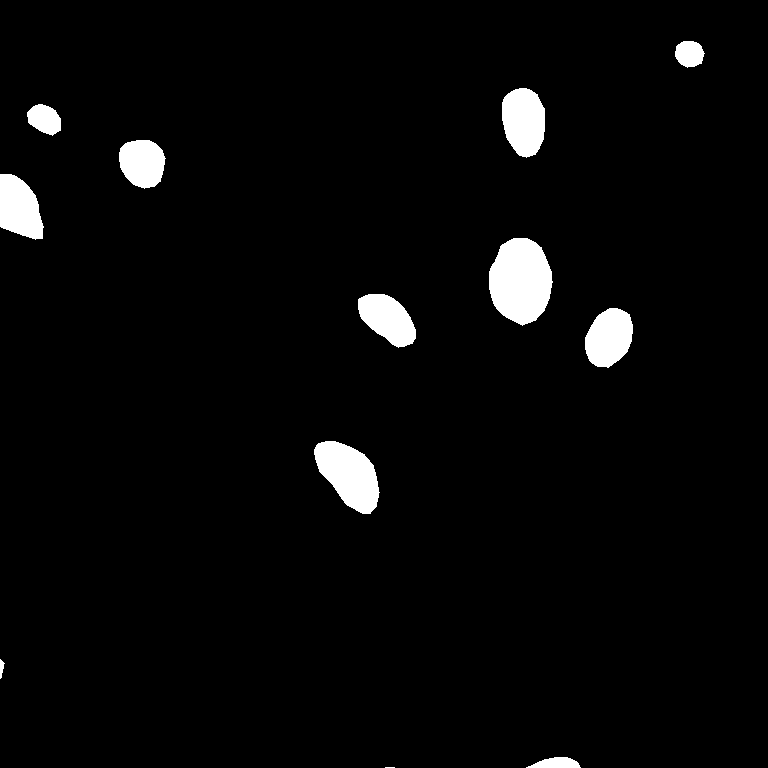}
        \caption{Lucchi mask}
    \end{subfigure}
    \begin{subfigure}[b]{.3\columnwidth}
        \centering
        \includegraphics[width=\linewidth]{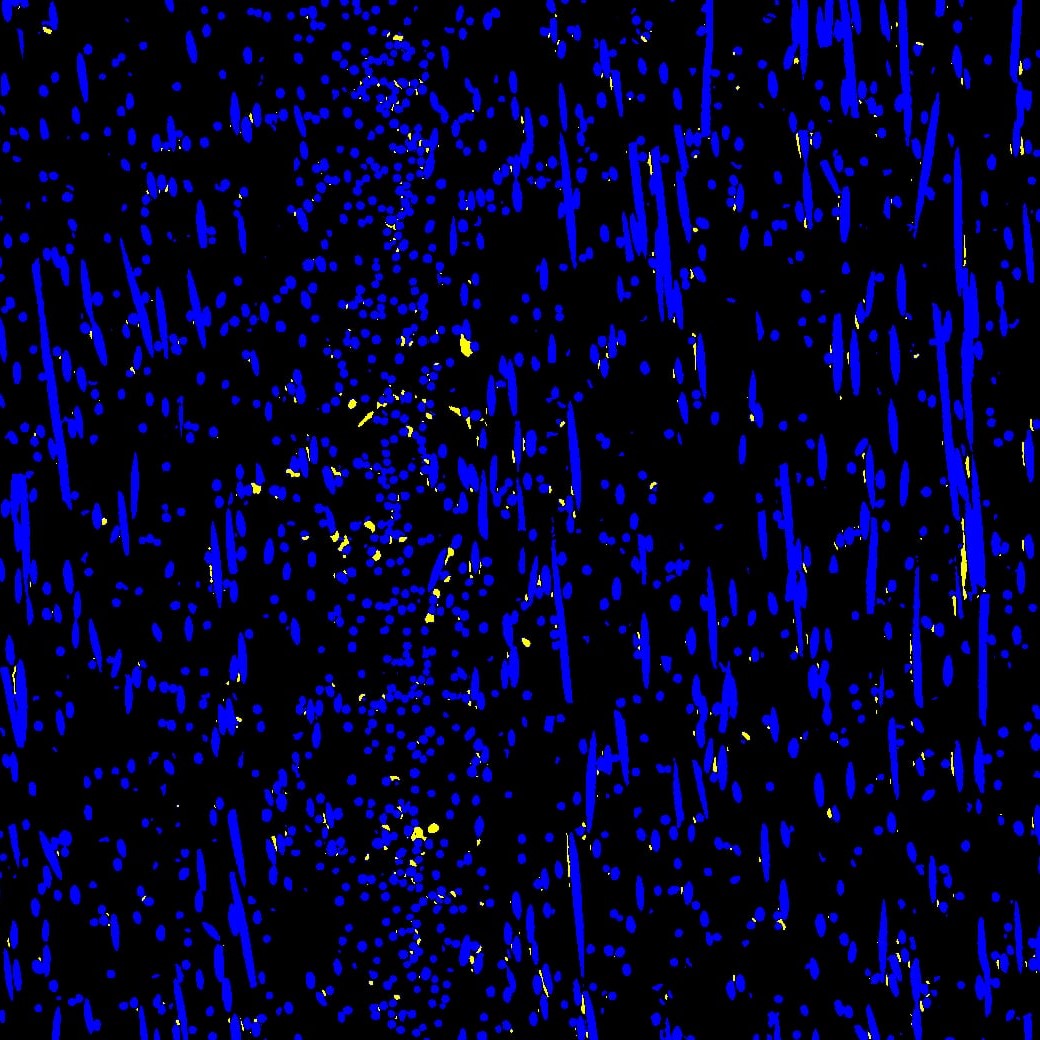}
        \caption{GF-PA66 mask}
    \end{subfigure}}
    \caption{
        Example images from the three different datasets used in the study. 
        The top row contains raw input images, while the bottom row shows their corresponding segmentation masks.
    }
    \label{fig:dataset_examples}
\end{figure}

To analyze the saliency asymmetry in 2D+ representations, experiments were conducted on three datasets: SNEMI, Lucchi, and GF-PA66. All datasets consist of grayscale images and were converted into three-channel inputs by stacking spatially adjacent slices following a 2D+ strategy. This setup enables models to incorporate local spatial context while allowing the saliency distributions at channel-level to be examined. Example randomly selected raw images and the corresponding segmentation masks are shown in \Cref{fig:dataset_examples}.

Although the primary focus of this study is on biomedical electron microscopy datasets, an additional non-biological dataset was included to assess the generalizability of the observed bias and to determine whether it is specific to biomedical imagery or also manifests in other grayscale imaging domains.

\subsection{SNEMI dataset}

The SNEMI dataset consists of high-resolution serial-section electron microscopy images of brain tissue and is widely used to benchmark neural circuit reconstruction tasks due to its annotation quality \cite{li2019dense, matejek2019biologically, gonda2021consistent, januszewski2019segmentation}. The grayscale slices depict fine neural structures, including membranes and synapses, with detailed ground truth annotations available for segmentation evaluation. However, it only consists of 100 samples with available ground truth label (75 train, 15 validation, 15 test), which makes it a great illustrative case for the importance of transfer-learning.

\subsection{Lucchi dataset}

The Lucchi dataset focuses on mitochondrial segmentation in brain tissue and contains anisotropic electron microscopy images with complex cellular environments. Subtle boundaries and variable contrast make this dataset particularly challenging, providing a sensitive test case to examine channel bias effects in pretrained models. The Lucchi dataset consists of 165 train and 165 test samples which are two sets of 2D recordings from different 3D slices.

\subsection{GF-PA66 3D XCT dataset}

The GF-PA66 3D XCT dataset comprises volumetric X-ray computed tomography scans of glass fiber-reinforced polyamide 66 composites. Unlike the other datasets, it captures structural variations and fiber orientation patterns of a composite material and includes two segmentation classes, enabling the evaluation of channel-level effects beyond biological imaging. Also, it consists of comparatively more image samples (1300 train, 300 test), however, they come from the same material sample with a 300 layer gap in between them to avoid information leakage.

Bertoldo \etal \cite{pa66} compared 2D, 2D+, and 3D configurations on this dataset using a U-Net architecture \cite{ronneberger2015u}, reporting comparable performance between 2D and 2D+ setups, and substantially lower performance for the 3D configuration. This outcome was attributed to data availability constraints affecting 3D training, nevertheless, the results highlight the potential advantages of 2D+ representations over fully 3D approaches when data or computational resources are limited. To ensure comparability with this prior work, the same architectures were adopted and focused on the 2D+ configuration in our study.

\section{Experimental Design} \label{sec:experiments}

The experiments were designed to assess the presence of saliency asymmetry, its consistency among attribution methods, the effectiveness of mitigation strategies, and its dependence on model architecture. All experimental configurations are summarized in \Cref{tab:experimental_setup}. The primary setup was applied across all datasets, with more detailed analyzes were conducted on the SNEMI dataset. Variations in dataset, architecture, encoder type, and weight initialization were used to isolate factors that influence channel bias in 2D+ representations.

\begin{table}[h!]
\centering
\resizebox{0.95\columnwidth}{!}{
\begin{tabular}{lcccl}
\hline
 & \textbf{Dataset} & \textbf{Model Type} & \textbf{Encoder Type} & \textbf{Pretrain Type} \\ \hline
\multirow{9}{*}{\begin{tabular}[c]{@{}l@{}}Saliency \\ Assymetry \\ Exictance\end{tabular}} & \multirow{3}{*}{SNEMI} & \multirow{3}{*}{DeepLabV3+} & \multirow{3}{*}{ResNet34} & ImageNet \\
 &  &  &  & Non-pretrained \\
 &  &  &  & Uniform-Green \\ \cline{2-5} 
 & \multirow{3}{*}{Lucchi} & \multirow{3}{*}{DeepLabV3+} & \multirow{3}{*}{ResNet34} & ImageNet \\
 &  &  &  & Non-pretrained \\
 &  &  &  & Uniform-Green \\ \cline{2-5} 
 & \multirow{3}{*}{GF-PA66} & \multirow{3}{*}{U-Net} & \multirow{3}{*}{ResNet34} & ImageNet \\
 &  &  &  & Non-pretrained \\
 &  &  &  & Uniform-Green \\ \hline
\multirow{7}{*}{\begin{tabular}[c]{@{}l@{}}Mitigation \\ Strategies\end{tabular}} & \multirow{7}{*}{SNEMI} & \multirow{7}{*}{DeepLabV3+} & \multirow{7}{*}{ResNet34} & Average \\
 &  &  &  & ImageNet \\
 &  &  &  & Lucchi-pretrained \\
 &  &  &  & Non-pretrained \\
 &  &  &  & Uniform-Blue \\
 &  &  &  & Uniform-Green \\
 &  &  &  & Uniform-Red \\ \hline
\multirow{4}{*}{\begin{tabular}[c]{@{}l@{}}Occurence in \\ different \\ calculation\end{tabular}} & \multirow{4}{*}{SNEMI} & \multirow{4}{*}{DeepLabV3+} & \multirow{4}{*}{ResNet34} & ImageNet \\
 &  &  &  & Lucchi-pretrained \\
 &  &  &  & Non-pretrained \\
 &  &  &  & Uniform-Green \\ \hline
\multirow{15}{*}{\begin{tabular}[c]{@{}l@{}}Model\\ Independance\end{tabular}} & \multirow{15}{*}{SNEMI} & \multirow{9}{*}{DeepLabV3+} & \multirow{3}{*}{ResNet18} & ImageNet \\
 &  &  &  & Non-pretrained \\
 &  &  &  & Uniform-Green \\ \cline{4-5} 
 &  &  & \multirow{3}{*}{ResNet34} & ImageNet \\
 &  &  &  & Non-pretrained \\
 &  &  &  & Uniform-Green \\ \cline{4-5} 
 &  &  & \multirow{3}{*}{ResNet50} & ImageNet \\
 &  &  &  & Non-pretrained \\
 &  &  &  & Uniform-Green \\ \cline{3-5} 
 &  & \multirow{3}{*}{SegFormer} & \multirow{3}{*}{ResNet34} & ImageNet \\
 &  &  &  & Non-pretrained \\
 &  &  &  & Uniform-Green \\ \cline{3-5} 
 &  & \multirow{3}{*}{U-Net} & \multirow{3}{*}{ResNet34} & ImageNet \\
 &  &  &  & Non-pretrained \\
 &  &  &  & Uniform-Green \\ \hline
\end{tabular}} 
\caption{Summary of experimental configurations}
\label{tab:experimental_setup}
\end{table}

\subsection{Model Architectures and Training Setup}

Three widely used 2D semantic segmentation architectures were selected: DeepLabV3+ \cite{chen2018encoder}, U-Net \cite{ronneberger2015u}, and SegFormer \cite{xie2021segformer}. These models are commonly used in 2D+ biomedical segmentation and represent distinct architectural paradigms. 

U-Net provides strong spatial localization through its encoder–decoder design with skip connections and remains a standard baseline in electron microscopy imaging. DeepLabV3+ extends convolutional architectures with atrous spatial pyramid pooling to capture multi-scale contextual information. SegFormer represents a modern transformer-based approach, using attention mechanisms to model global context efficiently and achieving state-of-the-art performance across many vision tasks. In transformer models attention plays a central role, as it directly governs how input tokens contribute to the output. Although attention modules are inherently permutation-invariant, positional embeddings introduce structured biases that make the ordering and identity of input tokens consequential. When applied to slice-stacked electron microscopy inputs, attention mechanisms pretrained on RGB data may therefore assign unequal importance to slices based on inherited channel and positional biases, rather than task-relevant structural cues. Together, the selected architectures enable systematic analysis of saliency asymmetry across convolutional and attention-based models within well-established and extensively studied biomedical imaging frameworks.

All models were implemented with ResNet-based \cite{he2016deep} encoders (ResNet18, ResNet34, ResNet50) to ensure comparability between experiments. The models were trained using binary cross-entropy with logits and optimized with Adam \cite{kingma2017adammethodstochasticoptimization}. The hyperparameters, including learning rate and weight decay, were empirically tuned to ensure stable convergence and high segmentation accuracy.

\subsection{Saliency Pattern Evaluation Setup}

Saliency maps were computed using the methods described in \Cref{sec:saliencymethods} to verify that observed asymmetries were not artifacts of a specific attribution technique. Although absolute attention distributions varied between methods, channel-level saliency patterns remained consistent, as illustrated in \Cref{fig:saliency_types} and further analyzed in \Cref{sec:occurencedifferentcalculation}. Models initialized with ImageNet pretrained weights, random initialization, and domain-specific pretraining were compared. Additional initialization strategies were explored, including channel-emphasized and averaged-channel initialization, to assess whether asymmetry could be reinforced or mitigated. This approach is conceptually related to the channel extrapolation strategies proposed by Messaoudi \etal \cite{messaoudi2023cross}, although their work focused on 3D models and did not analyze channel-level attention.

To achieve the proposed mitigation strategies, modifications to the initial weights of the three channels derived from ImageNet pretraining were experimented with. In each case, the pretraining weights of one channel were selected and applied uniformly to all three channels. These models are denoted as "uniform" followed by the selected channel's name (e.g., "Uniform-Green" refers to all channels being initialized with the ImageNet pretrained weights of the green channel). Additionally, an approach in which the initial weights were set to the average of the three ImageNet pretrained channels, referred to as the "Average" initialization, was tested. Finally, a transfer learning strategy was explored in which a model was first trained using the "Uniform-Green" initialization on the Lucchi dataset, and its learned weights were subsequently used to initialize the final model on SNEMI.

\subsection{Channel Attention Difference Measures}

Channel-level differences in saliency distributions were quantified using Jensen--Shannon Entropy \cite{lin1991divergence, kullback1951information}, Bhattacharyya Distance \cite{bhattacharyya1943measure, bhattacharyya1946measure}, and Wasserstein Distance \cite{kantorovich1942translocation}. While Jensen-Shannon and Bhattacharyya distances capture point-wise distribution differences, Wasserstein Distance (Earth Mover’s Distance) measures the minimum cost of transforming one distribution into another and provides a more continuous and interpretable metric. Although all three measures exhibited similar trends, Wasserstein Distance was selected for primary analysis due to its sensitivity to relative distribution proximity. Jensen-Shannon and Bhattacharyya results are provided in the supplementary material.

In addition, two Wasserstein-based metrics were defined. The Symmetric Wasserstein Distance (\textit{SWd}) measures symmetry between equidistant channels while ignoring overall magnitude differences, which may naturally arise due to the central prediction slice. The Full Wasserstein Distance (\textit{FWd}) captures the overall distribution differences between channels. Although a high \textit{FWd} alone does not indicate bias, a high \textit{SWd} reflects true asymmetry. Consequently, \textit{SWd} is used as the primary bias measure, with \textit{FWd} reported for context. Formally, if $W(d_{n_1}, d_{n_2})$ denotes the Wasserstein Distance between two distributions, $N$ is the number of channels (which should always be an odd number), and $Sd_n$ represents the distribution of saliency intensity values for channel $n\in[0,\dots,N\text{-}1]$, then \textit{SWd} and \textit{FWd} are defined as follows:

\begin{equation}
    SWd = \sum_{n \in \{1, \dots, \frac{(N\text{-}1)}{2}\}} 
    \frac{W \left( Sd_{-n + \frac{(N\text{-}1)}{2}}, Sd_{n + \frac{(N\text{-}1)}{2}} \right)}{\frac{(N\text{-}1)}{2}}
\end{equation}

\begin{equation}
    FWd = \sum_{\substack{0 \leq n < m < N}} 
    \frac{W(Sd_{n}, Sd_{m})}{\frac{N(N\text{-}1)}{2}}
\end{equation}

\section{Results}

This section presents empirical evidence of saliency asymmetry in slice-stacked electron microscopy inputs and evaluates targeted strategies for mitigating this effect. In addition to saliency-based analyzes, segmentation performances were reported to verify that all saliency evaluations are conducted on well-performing models and to examine whether mitigation strategies affect predictive quality. The analysis spans multiple datasets, model architectures, encoder backbones, and saliency formulations, demonstrating that the observed asymmetry is a systematic property of pretrained color-channel initialization rather than an artifact of a specific model or explanation method. All experimental configurations are summarized in \Cref{tab:experimental_setup}.

Unless stated otherwise, experiments are conducted on the SNEMI dataset, which serves as the primary benchmark due to its larger sample size. Baseline experiments employ DeepLabV3+ with a ResNet34 encoder, trained using ImageNet-pretrained, randomly initialized (Non-pretrained), and Uniform-Green initialized weights. Saliency maps are computed using both foreground-only and full-output formulations, and saliency asymmetry is quantified primarily using the Wasserstein Distance, with symmetric distances between the two side channels (Channel1 and Channel3) treated as the main indicator of channel imbalance. For baseline predictive performance metric Dice score was selected.

\subsection{Segmentation Performances}
\begin{figure}[h!]
\centering
\resizebox{0.95\columnwidth}{!}{
    \centering
    \begin{subfigure}[b]{.32\columnwidth}
        \centering
        \includegraphics[width=\linewidth]{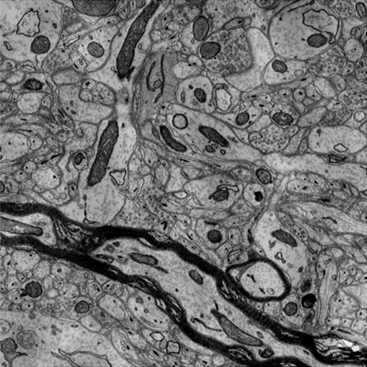}
        \caption{Original Image}
    \end{subfigure}
    \begin{subfigure}[b]{.32\columnwidth}
        \centering
        \includegraphics[width=\linewidth]{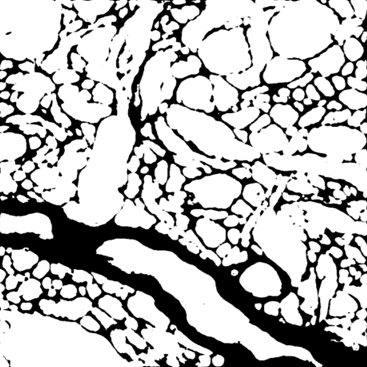}
        \caption{Predicted Mask}
    \end{subfigure}
    \begin{subfigure}[b]{.32\columnwidth}
        \centering
        \includegraphics[width=\linewidth]{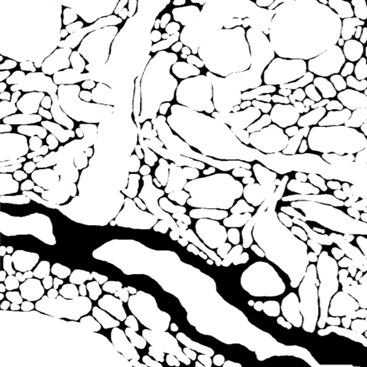}
        \caption{Ground Truth Mask}
    \end{subfigure}}
    \caption{Example prediction result for the SNEMI dataset. The predicted mask is thresholded at the confidence value of 0.85.}
\label{fig:snemi_prediction_example}
\end{figure}

Before analyzing saliency distributions, it was verified that all models produced accurate and stable segmentation outputs. This step is essential, as saliency analysis is only meaningful when applied to predictions of comparable quality: models with similar performance may rely on different internal representations, and such differences are precisely the focus of the present study. Model performance was assessed quantitatively using Dice score, IoU, Precision, Recall, Accuracy, and BCE-based loss, complemented by qualitative inspection. A representative SNEMI example and its corresponding prediction are shown in \Cref{fig:snemi_prediction_example}. The example illustrates that the evaluated models achieve satisfactory segmentation quality, supporting the validity of the subsequent saliency analyses.

\vspace{4pt}

\begin{table*}[h!]
\centering
\resizebox{0.8\textwidth}{!}{ 
\begin{tabular}{llllllll}
\hline
\multicolumn{1}{c}{\textbf{ModelType}} & \multicolumn{1}{c}{\textbf{Encoder}} & \multicolumn{1}{c}{\textbf{PretrainStatus}} & \multicolumn{1}{c}{\textbf{Dice (± SE)}} & \multicolumn{1}{c}{\textbf{IoU (± SE)}} & \multicolumn{1}{c}{\textbf{Precision (± SE)}} & \multicolumn{1}{c}{\textbf{Recall (± SE)}} & \multicolumn{1}{c}{\textbf{Accuracy (± SE)}} \\ \hline
\multirow{19}{*}{DeepLabV3+} & \multirow{6}{*}{ResNet18} & Average & 0.9329 ± 0.001 & 0.8743 ± 0.0018 & 0.8791 ± 0.0021 & 0.9938 ± 0.0004 & 0.8845 ± 0.0019 \\
 &  & ImageNet & 0.9331 ± 0.0012 & 0.8747 ± 0.0021 & 0.8824 ± 0.0021 & 0.9902 ± 0.0012 & 0.8853 ± 0.0021 \\
 &  & Non-pretrained & \textit{0.9295 ± 0.001} & \textit{0.8683 ± 0.0017} & \textit{0.8753 ± 0.0016} & 0.991 ± 0.0013 & \textit{0.8786 ± 0.0017} \\
 &  & Uniform-Blue & \textbf{0.9376 ± 0.0011} & \textbf{0.8825 ± 0.0019} & \textbf{0.8931 ± 0.002} & \textit{0.9867 ± 0.0006} & \textbf{0.8938 ± 0.0019} \\
 &  & Uniform-Green & 0.9321 ± 0.0011 & 0.8729 ± 0.0019 & 0.8773 ± 0.0022 & \textbf{0.9943 ± 0.0004} & 0.883 ± 0.002 \\
 &  & Uniform-Red & 0.9352 ± 0.0009 & 0.8782 ± 0.0016 & 0.885 ± 0.0015 & 0.9913 ± 0.0005 & 0.8889 ± 0.0016 \\ \cline{2-8} 
 & \multirow{7}{*}{ResNet34}  & Average & 0.9344 ± 0.0011 & 0.8769 ± 0.0019 & 0.8829 ± 0.0019 & 0.9923 ± 0.0005 & \textit{0.8874 ± 0.0019} \\
 &  & ImageNet & 0.9363 ± 0.0012 & 0.8803 ± 0.0021 & 0.8871 ± 0.0025 & 0.9914 ± 0.0006 & 0.8909 ± 0.0022 \\
 &  & LucchiPretrained & 0.936 ± 0.0012 & 0.8798 ± 0.002 & 0.8873 ± 0.0021 & 0.9905 ± 0.0004 & 0.8907 ± 0.002 \\
 &  & Non-pretrained & \textit{0.9281 ± 0.0009} & \textit{0.866 ± 0.0016} & \textit{0.8722 ± 0.0015} & 0.9918 ± 0.0007 & 0.876 ± 0.0016 \\
 &  & Uniform-Blue & 0.9388 ± 0.0011 & 0.8846 ± 0.0019 & \textbf{0.8946 ± 0.0022} & \textit{0.9875 ± 0.0004} & 0.8959 ± 0.002 \\
 &  & Uniform-Green & 0.935 ± 0.0011 & 0.878 ± 0.002 & 0.8841 ± 0.0024 & \textbf{0.9924 ± 0.0006} & 0.8885 ± 0.0021 \\
 &  & Uniform-Red & \textbf{0.939 ± 0.001} & \textbf{0.885 ± 0.0018} & 0.893 ± 0.0018 & 0.9899 ± 0.0003 & \textbf{0.896 ± 0.0017} \\ \cline{2-8} 
 & \multirow{6}{*}{ResNet50} & Average & 0.9358 ± 0.0014 & 0.8795 ± 0.0025 & 0.8871 ± 0.0033 & 0.9906 ± 0.0011 & 0.8901 ± 0.0027 \\
 &  & ImageNet & 0.9372 ± 0.0011 & 0.8819 ± 0.002 & 0.8892 ± 0.0023 & 0.9908 ± 0.0004 & 0.8927 ± 0.0021 \\
 &  & Non-pretrained & \textit{0.9256 ± 0.0008} & \textit{0.8616 ± 0.0014} & \textit{0.8697 ± 0.001} & 0.9892 ± 0.0014 & \textit{0.8716 ± 0.0013} \\
 &  & Uniform-Blue & \textbf{0.9405 ± 0.001} & \textbf{0.8877 ± 0.0019} & \textbf{0.898 ± 0.0021} & \textit{0.9872 ± 0.0005} & \textbf{0.899 ± 0.0019} \\
 &  & Uniform-Green & 0.9359 ± 0.0011 & 0.8796 ± 0.0019 & 0.8863 ± 0.0022 & \textbf{0.9916 ± 0.0005} & 0.8903 ± 0.002 \\
 &  & Uniform-Red & 0.9401 ± 0.0012 & 0.887 ± 0.0021 & 0.8938 ± 0.0021 & \textbf{0.9916 ± 0.0002} & 0.898 ± 0.002 \\ \hline
\multirow{3}{*}{SegFormer} & \multirow{3}{*}{ResNet34} & ImageNet & \textbf{0.9415 ± 0.0011} & \textbf{0.8896 ± 0.0019} & \textbf{0.899 ± 0.0022} & \textit{0.9883 ± 0.0007} & \textbf{0.9008 ± 0.0019} \\
 &  & Non-pretrained & \textit{0.9313 ± 0.0009} & \textit{0.8715 ± 0.0016} & \textit{0.8779 ± 0.0018} & \textbf{0.9917 ± 0.0014} & \textit{0.8818 ± 0.0015} \\
 &  & Uniform-Green & 0.9397 ± 0.0011 & 0.8863 ± 0.0019 & 0.8933 ± 0.002 & 0.9913 ± 0.0003 & 0.8973 ± 0.0019 \\ \hline
\multirow{3}{*}{U-Net} & \multirow{3}{*}{ResNet34} & ImageNet & \textbf{0.9429 ± 0.0011} & \textbf{0.892 ± 0.002} & \textbf{0.8996 ± 0.0022} & 0.9906 ± 0.0004 & \textbf{0.9031 ± 0.002} \\
 &  & Non-pretrained & \textit{0.929 ± 0.0013} & \textit{0.8674 ± 0.0023} & \textit{0.8737 ± 0.0021} & \textit{0.992 ± 0.0031} & \textit{0.8776 ± 0.0021} \\ 
 &  & Uniform-Green & 0.9405 ± 0.0012 & 0.8877 ± 0.0021 & 0.8937 ± 0.0023 & \textbf{0.9925 ± 0.0004} & 0.8985 ± 0.0021 \\ \hline
\end{tabular}}
\caption{A complete list of performance evaluation results for the semantic segmentation task on the SNEMI dataset. Different pretraining statuses yield nearly identical results within the margin of error, while Non-pretrained models almost always show lower performance.}
\label{tab:snemi_all_pretrain_performance}
\end{table*}

\vspace{4pt}

Across all tested architectures and encoders, Non-pretrained models consistently underperformed relative to all pretrained variants. This difference was statistically significant between ImageNet-pretrained and Uniform-Green techniques and was presented based on Dice score (ImageNet vs Non-pretrained: Mann–Whitney U, $p = 1.3751*10^{-20}$, Cliff’s  $\delta$ $= 0.6949$, large effect; Non-pretrained vs Uniform-Green: Mann–Whitney U, $p = 7.1458*10^{-17}$; Cliff’s $\delta$ $= -0.6233$, large effect), confirming that some form of structured initialization is necessary for competitive segmentation performance. In contrast, ImageNet-pretrained and uniform-channel initialization strategies yielded nearly identical Dice, IoU, and Accuracy scores within the reported standard errors (\Cref{tab:snemi_all_pretrain_performance}). Pairwise comparisons between ImageNet and uniform strategies (Uniform-Green selected as baseline) using the Mann–Whitney U test showed that the difference in Dice and IoU was generally not statistically significant (Dice: $p = 0.0741$, Cliff’s $\delta$ $= 0.1335$, negligible effect; IoU: $p = 0.0741$, Cliff’s $\delta$  $= 0.1335$, negligible effect]), supporting the conclusion of comparable overlap-based segmentation quality. The most pronounced differences between the pretrained variants appeared in the Precision–Recall trade-off: Uniform-Blue and Uniform-Green initializations tended to favor recall at the expense of precision, whereas ImageNet-pretrained models exhibited slightly higher precision with lower recall. Crucially, the recall advantage of uniform initializations was statistically significant (ImageNet vs Uniform-Green: Mann–Whitney U, $p = 1.1314*10^{-06}$, Cliff’s $\delta$ $= -0.3636$, medium effect), while the precision advantage of ImageNet, when present, was typically not widely supported (ImageNet vs Uniform-Green: Mann–Whitney U, $p = 0.0122$, Cliff’s $\delta$ $= 0.1874$, small effect). These results establish that the mitigation strategies investigated in the following sections preserve segmentation quality while allowing a meaningful comparison of saliency behavior.

\subsection{The existence of Saliency Asymmetry in Pretrained Models}

\begin{figure}[h!]
\centering
\resizebox{0.95\columnwidth}{!}{
    \centering
    \begin{subfigure}[b]{\columnwidth}
        \centering
        \includegraphics[width=\linewidth]{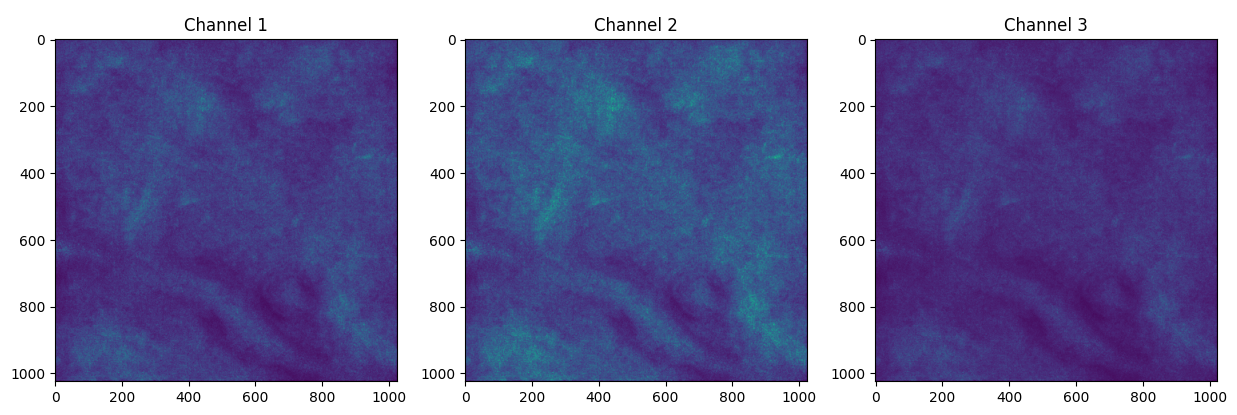}
        \subcaption{Averaged Saliency Map for resnet34 - ImageNet}
    \end{subfigure}}
\centering
\resizebox{0.95\columnwidth}{!}{
    \centering
    \begin{subfigure}[b]{\columnwidth}
        \centering
        \includegraphics[width=\linewidth]{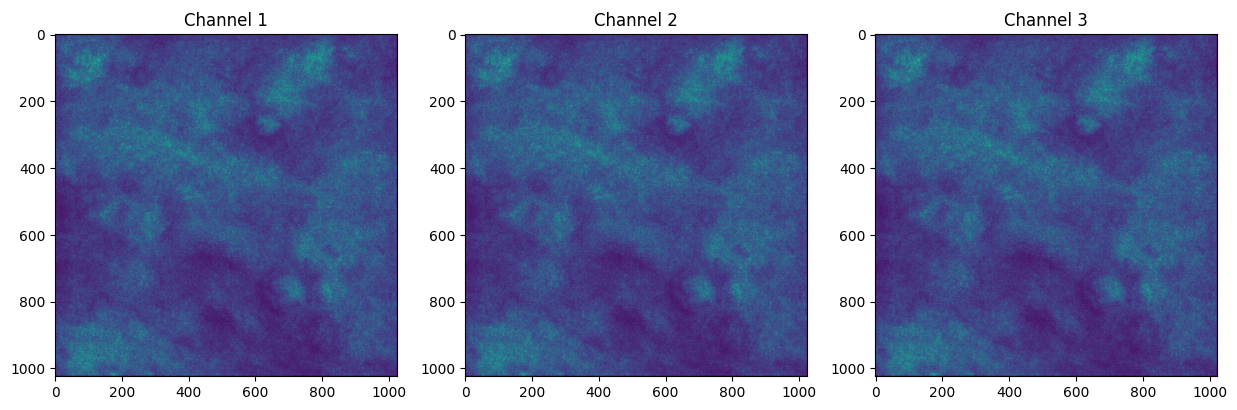}
        \subcaption{Averaged Saliency Map for resnet34 - Uniform-Green}
    \end{subfigure}}
        
    \caption{SNEMI dataset example of averaged "Foreground" saliency map calculation using ImageNet and "Uniform-Green" pretrained models, showing highly variable channel attention in ImageNet, while showing almost no variance in "Uniform-Green".}
    \label{fig:saliency_plot}
\end{figure}

An example of averaged foreground saliency maps is shown in \Cref{fig:saliency_plot}. When ImageNet-pretrained weights are used, saliency is unevenly distributed across channels, with Channel2 and Channel1 receiving substantially more attention than Channel3. In contrast, the Uniform-Greenn initialized model exhibits a markedly more balanced channel-wise saliency distribution.

\vspace{4pt}

\Cref{fig:top_models} confirms the existence of saliency asymmetry in models trained with pretrained color channel weights in multiple datasets, including SNEMI, Lucchi, and the GF-PA66 3D XCT dataset. This asymmetry is particularly evident when using ImageNet-pretrained weights, where models display an inherent bias toward certain color channels. However, this bias is not exclusive to the ImageNet-pretrained models, as in some cases even Non-pretrained models showed similar imbalances.

\begin{table}[h!]
\centering
\resizebox{0.95\columnwidth}{!}{
\begin{tabular}{llcccc}
\hline
 & \multicolumn{1}{c}{\textbf{}} & \multicolumn{2}{c}{\textbf{\begin{tabular}[c]{@{}c@{}}Symmetric \\ Wasserstein Distance\end{tabular}}} & \multicolumn{2}{c}{\textbf{\begin{tabular}[c]{@{}c@{}}Full \\ Wasserstein Distance\end{tabular}}} \\ \cline{3-6} 
 &  & Foreground & Full Output & Foreground & Full Output \\  \hline 
\multirow{3}{*}{\textbf{SNEMI}} & ImageNet & \textit{30.555} & \textit{32.060} & \textit{57.417} & \textit{60.164} \\
 & Non-pretrained & 9.585 & 9.433 & 11.747 & 11.275 \\
 & Uniform-Green & \textbf{0.282} & \textbf{0.249} & \textbf{10.347} & \textbf{8.767} \\ \hline
\multirow{3}{*}{\textbf{Lucchi}} & ImageNet & \textit{15.517} & \textit{27.488} & \textit{19.091} & \textit{33.485} \\
 & Non-pretrained & \textbf{0.214}  & 2.199 & 1.237 & 4.544 \\
 & Uniform-Green & 0.462 & \textbf{1.269} & \textbf{0.468} & \textbf{1.271} \\ \hline
\multirow{3}{*}{\textbf{GF-PA66}} & ImageNet & \textit{41.361} & \textit{47.205} & 114.209 & 109.641 \\
 & Non-pretrained & 5.967 & 5.092 & \textit{132.477} & \textit{133.590} \\
 & Uniform-Green & \textbf{0.375} & \textbf{0.367} & \textbf{99.018} & \textbf{85.950} \\ \hline
\end{tabular}}
\caption{Demonstration of the existence and extent of saliency asymmetry in pretrained models based on Symmetric and Full Wasserstein Distance.}
\label{fig:top_models}
\end{table}

This qualitative observation is quantitatively confirmed in \Cref{fig:top_models}, which reports Symmetric and Full Wasserstein Distances across multiple datasets, including SNEMI, Lucchi, and the GF-PA66 3D XCT dataset. ImageNet-pretrained models consistently exhibit the highest degree of saliency asymmetry, particularly in the Symmetric Wasserstein Distance between the two side channels. Non-pretrained models generally display reduced asymmetry compared to ImageNet-pretrained models, though notable imbalances remain in several cases. Nonparametric statistical testing confirms that these differences are not attributable to random variation. Subsequent pairwise Mann–Whitney U tests showed that ImageNet-pretrained models differ significantly from Non-pretrained and uniform-initialized models, with large effect sizes (Cliff’s $\delta$ ranges from approximately $0.5$ to $1.0$, depending on dataset and the formulation of the saliency; for all dataset ImageNet vs Non-pretrained: $p = 3.3902*10^{-09}$, Cliff’s $\delta$ $= 0.5283$, large effect; ImageNet vs Uniform-Green: $p = 1.7604*10^{-18}$, Cliff’s $\delta$ $= 0.7666$, large effect). These results provide strong evidence that saliency asymmetry is an inherent and systematic property of ImageNet-pretrained color-channel initialization.
An exception is observed for the full Wasserstein Distance on the GF-PA66 dataset, where the Non-pretrained model exhibits a larger overall distance than the ImageNet-pretrained model. Closer inspection reveals that this effect is driven primarily by deviations in the middle channel, whereas the side channels remain relatively symmetric. Since the present work focuses on the symmetry of contextual information provided by the side channels, the Symmetric Wasserstein Distance is considered the more relevant metric. This choice is consistent with biomedical assumptions, where the middle slice serves as the primary source of predictive information and adjacent slices provide contextual support.

\subsection{Mitigation Strategies for Saliency Asymmetry}

Given the magnitude and consistency of the observed saliency asymmetry, mitigation strategies were developed with the goal of reducing channel imbalance while preserving the benefits of pretraining. Rather than completely eliminating pretraining, modifications of the initial ImageNet-derived weights were explored.

Specifically, uniform-channel initializations were constructed by copying the pretrained weights of a single ImageNet color channel (red, green, or blue) across all three input channels. In addition, an Average initialization was tested, in which the mean of the three pretrained channels was used. Finally, a multi-stage transfer learning variant was evaluated in which a Uniform-Green–initialized model was first trained on the Lucchi dataset and then used to initialize the target model in SNEMI.

The corresponding results of the saliency asymmetry are summarized in \Cref{tab:mitigations}. ImageNet-pretrained models exhibit substantially higher Symmetric and Full Wasserstein Distances than all mitigation strategies. Uniform-channel initializations dramatically reduce saliency asymmetry. Among the tested approaches, the Average initialization consistently performs the worst in terms of asymmetry reduction. Uniform-Green and Lucchi-pretrained initializations yield the most balanced saliency distributions, a result that aligns with the observation that the green channel in natural images typically represents broadly distributed intensity information rather than semantically localized structures.
Pairwise statistical comparisons confirm that the reduction in saliency asymmetry achieved by uniform-channel strategies is significant relative to ImageNet pretraining (ImageNet vs Uniform-Green: $p = 1.7604*10^{-18}$, Cliff’s $\delta$ $= 0.7666$, large effect), with very large effect sizes (Cliff’s $\delta \approx -0.7$ to $-0.9$). Notably, uniform-channel models also exhibit lower asymmetry than Non-pretrained models, but not as much as ImageNet (Non-pretrained vs UniformGreen: $p = 9.5091*10^{-08}$, Cliff’s $\delta$ $= 0.4663$, medium effect).

\begin{table}[h!]
\centering
\resizebox{0.95\columnwidth}{!}{
\begin{tabular}{lcccc}
\hline
 & \multicolumn{2}{c}{\textbf{\begin{tabular}[c]{@{}c@{}}Symmetric\\ Wasserstein Distance\end{tabular}}} & \multicolumn{2}{c}{\textbf{\begin{tabular}[c]{@{}c@{}}Full\\ Wasserstein Distance\end{tabular}}} \\ \cline{2-5} 
 & Foreground & Full Output & Foreground & Full Output \\ \hline
Average & 0.053 & 0.047 & 20.822 & 19.126 \\
ImageNet & \textit{30.555} & \textit{32.060} & \textit{57.417} & \textit{60.164} \\
Lucchi Pretrained & \textbf{0.061} & \textbf{0.059} & \textbf{7.091} & 6.336 \\
Non-pretrained & 9.585 & 9.433 & 11.747 & 11.275 \\
Uniform-Blue & 0.0390 & 0.021 & 10.285 & \textbf{5.589} \\
Uniform-Green & 0.282 & 0.249 & 10.347 & 8.767 \\
Uniform-Red & 0.147 & 0.140 & 8.308 & 8.013 \\ \hline
\end{tabular}}
\caption{Mitigation strategies for saliency asymmetry based on Symmetric and Full Wasserstein Distance, using "Foreground" and "Full Output" summation saliency map generation strategies. Top performers are marked in bold, while bottom performers are marked in italic.}
\label{tab:mitigations}
\end{table}

\subsection{Persistance of Saliency Asymmetry Across Different Saliency Calculation Methods}  \label{sec:occurencedifferentcalculation}

To assess whether saliency asymmetry depends on the specific explanation technique, multiple saliency map calculation methods were evaluated, including foreground-only and full-output summation strategies, GradCAM++, and occlusion-based saliency.
As shown in \Cref{tab:differentsaliencycalculation}, ImageNet-pretrained models consistently exhibit the highest Symmetric and Full Wasserstein Distances in all saliency formulations. Non-pretrained models display intermediate asymmetry, while uniform-channel and Lucchi-pretrained models maintain low Wasserstein Distances across methods. Although absolute distance values vary substantially between saliency techniques, the relative order of pretraining strategies remains stable.
These results demonstrate that saliency asymmetry is not an artifact of a particular explanation method. Instead, it reflects a persistent property of the learned representations induced by pretrained color-channel initialization.

\begin{table}[h!]
\centering
\resizebox{0.95\columnwidth}{!}{
\begin{tabular}{lcccccc}
\hline
 & \multicolumn{6}{c}{\textbf{Symmetric Wasserstein Distance}} \\ \cline{2-7} 
 & Foreground & \begin{tabular}[c]{@{}c@{}}Foreground\\ 100\end{tabular} & Full Output & \begin{tabular}[c]{@{}c@{}}Full Output\\ 100\end{tabular} & GradCAM++ & Occlusion \\ \hline
ImageNet & \textit{30.555} & \textit{14.754} & \textit{32.060} & \textit{19.475} & \textit{183.028} & \textit{22.243} \\
Lucchi Pretrained & \textbf{0.061} & \textbf{0.020} & \textbf{0.059} & \textbf{0.015} & 70.072 & 10.308 \\
Non-pretrained & 9.585 & 2.883 & 9.433 & 2.967 & 68.798 & \textbf{8.519} \\
Uniform-Green & 0.282 & 0.083 & 0.249 & 0.145 & \textbf{22.521} & 9.750 \\ \hline
 & \multicolumn{6}{c}{\textbf{Full Wasserstein Distance}} \\ \cline{2-7} 
 & Foreground & \begin{tabular}[c]{@{}c@{}}Foreground\\ 100\end{tabular} & Full Output & \begin{tabular}[c]{@{}c@{}}Full Output\\ 100\end{tabular} & GradCAM++ & Occlusion \\ \hline
ImageNet & \textit{57.417} & \textit{27.413} & \textit{60.164} & \textit{36.074} & \textit{122.589} & \textit{89.418} \\
Lucchi Pretrained & \textbf{7.091} & \textbf{2.994} & \textbf{6.336} & \textbf{2.335} & 46.715 & \textbf{28.605} \\
Non-pretrained & 11.747 & 3.376 & 11.275 & 3.538 & 69.366 & 49.458 \\
Uniform-Green & 10.347 & 3.038 & 8.767 & 5.107 & \textbf{19.869} & 47.930 \\ \hline
\end{tabular}}
\caption{Demonstration of the persistence of saliency asymmetry across different saliency map calculation methods, based on Symmetric and Full Wasserstein Distance. Top performers are marked in bold, while bottom performers are marked in italic.}
\label{tab:differentsaliencycalculation}
\end{table}

\subsection{Model-Independance of Saliency Asymmetry}

Finally, it was evaluated whether the saliency asymmetry depends on the model architecture or the complexity of the encoder. Experiments were conducted using DeepLabV3+, SegFormer, and U-Net architectures, as well as multiple ResNet backbones within the DeepLabV3+ framework.
The results, summarized in \Cref{tab:differentmodelssaliency}, consistently mirror the patterns observed in earlier sections. Across all architectures and backbones, ImageNet-pretrained models exhibit pronounced saliency asymmetry, whereas Uniform-Green–initialized models maintain near-symmetric channel-wise saliency distributions. Non-pretrained models again occupy an intermediate position.
These findings confirm that the saliency asymmetry is largely independent of architectural choice and encoder depth. Instead, it arises primarily from pretrained color-channel initialization and can be effectively mitigated through uniform-channel strategies without sacrificing segmentation performance.

\begin{table}[h!]
    \centering
    \begin{subtable}{0.95\columnwidth}
        \centering
        \resizebox{\columnwidth}{!}{
        \begin{tabular}{llcccc}
            \hline
 &  & \multicolumn{2}{c}{\textbf{\begin{tabular}[c]{@{}c@{}}Symmetric\\ Wasserstein Distance\end{tabular}}} & \multicolumn{2}{c}{\textbf{\begin{tabular}[c]{@{}c@{}}Full \\ Wasserstein Distance\end{tabular}}} \\ \cline{3-6} 
 &  & Full Output & Full Output Mask & Full Output & Full Output Mask \\ \hline
            \multirow{3}{*}{\textbf{DeepLabV3+}} & ImageNet & \textit{30.555} & \textit{32.060} & \textit{57.417} & \textit{60.164} \\
            & Non-pretrained & 9.585 & 9.433 & 11.747 & 11.275 \\
            & Uniform-Green & \textbf{0.282} & \textbf{0.249} & \textbf{10.347} & \textbf{8.767} \\ \hline
            \multirow{3}{*}{\textbf{SegFormer}} & ImageNet & \textit{38.425} & \textit{38.674} & \textit{70.052} & \textit{70.487} \\
            & Non-pretrained & 1.236 & 1.374 & \textbf{3.472} & \textbf{3.729} \\
            & Uniform-Green & \textbf{0.234} & \textbf{0.268} & 11.542 & 13.197 \\ \hline
            \multirow{3}{*}{\textbf{U-Net}} & ImageNet & \textit{21.514} & \textit{19.985} & \textit{40.476} & \textit{37.578} \\
            & Non-pretrained & 3.088 & 2.611 & \textbf{8.376} & \textbf{6.433} \\
            & Uniform-Green & \textbf{0.220} & \textbf{0.154} & 12.347 & 8.283 \\ \hline
        \end{tabular}}
        \caption{Comparison of DeepLabV3+, SegFormer, and U-Net models.}
    \end{subtable}
    
    \vspace{0.2cm} 
    
    \begin{subtable}{0.95\columnwidth}
        \centering
        \resizebox{\columnwidth}{!}{
        \begin{tabular}{llcccc}
            \hline
 &  & \multicolumn{2}{c}{\textbf{\begin{tabular}[c]{@{}c@{}}Symmetric\\ Wasserstein Distance\end{tabular}}} & \multicolumn{2}{c}{\textbf{\begin{tabular}[c]{@{}c@{}}Full \\ Wasserstein Distance\end{tabular}}} \\ \cline{3-6} 
 &  & Full Output & Full Output Mask & Full Output & Full Output Mask \\ \hline
\multirow{3}{*}{\textbf{ResNet18}} & ImageNet & \textit{45.063} & \textit{47.861} & \textit{80.922} & \textit{85.863} \\
 & Non-pretrained & 7.630 & 6.592 & 18.078 & 16.520 \\
 & Uniform-Green & \textbf{0.135} & \textbf{0.191} & \textbf{6.937} & \textbf{9.037} \\ \hline
\multirow{3}{*}{\textbf{ResNet34}} & ImageNet & \textit{32.060} & \textit{30.555} & \textit{60.164} & \textit{57.417} \\
 & Non-pretrained & 9.433 & 9.585 & 11.275 & 11.747 \\
 & Uniform-Green & \textbf{0.249} & \textbf{0.282} & \textbf{8.767} & \textbf{10.347} \\ \hline
\multirow{3}{*}{\textbf{ResNet50}} & ImageNet & \textit{23.667} & \textit{31.100} & \textit{48.859} & \textit{64.151} \\
 & Non-pretrained & 10.234 & 9.026 & \textbf{6.823} & \textbf{6.017} \\
 & Uniform-Green & \textbf{0.144} & \textbf{0.188} & 9.051 & 11.622 \\ \hline
        \end{tabular}}
        \caption{Comparison of ResNet18, ResNet34, and ResNet50 backbones.}
    \end{subtable}
\caption{Demonstration of the model independence of saliency asymmetry for both entirely different model architectures (a) and the same DeepLabV3+ architecture with varying encoder backbone complexities (b). Top performers are marked in bold, while bottom performers are marked in italic.}
\label{tab:differentmodelssaliency}
\end{table}

\section{Discussion}

Our experiments demonstrate that pretrained model weights substantially influence saliency map distributions, particularly in specialized datasets such as biomedical electron microscopy and X-ray Computed Tomography (XCT) images. Transferring color channel weights from natural image datasets, such as ImageNet, introduces inherent asymmetry in saliency maps. This bias arises from the color distribution patterns in natural images and manifests as preferential weighting of specific channels, leading to unnatural model attention in domains with distinct channel characteristics. Although biomedical datasets constituted the primary focus of this study due to prior domain knowledge, the results from the GF-PA66 dataset suggest that the observed phenomenon is not restricted to biomedical imaging and may generalize to other grayscale imaging domains.

Saliency asymmetry persists across multiple model architectures --- both convolutional and transformer-like model --- and encoder backbones, highlighting that pretrained weights can introduce systematic biases even when transfer learning improves predictive performance. These biases compromise the interpretability of channel attention, creating dependencies that do not correspond to the actual structural or biological features of the data. Consequently, explanations derived from attention maps in such models may be unreliable.

To mitigate this effect, a simple yet effective modification was evaluated, initializing all input channels with the pretrained green channel from ImageNet (“Uniform-Green”). This strategy substantially reduced asymmetry, producing more balanced saliency distributions while preserving the benefits of pretrained weights. Alternative approaches, such as averaging channel weights or domain-specific pretraining (Lucchi-pretrained model), also reduced bias. Furthermore, the findings suggest that the pretrained weights generated through domain-specific training could serve as reusable initializations, similar to the use of Lucchi-pretrained model in this work. The public release of such adapted weights supports researchers working with similar 2D+ data formats, providing improved initialization strategies without the limitations of natural image-based pretraining.

\section{Conclusion}

This study demonstrates that pretrained weights from natural image datasets introduce systematic attention bias in specialized image segmentation tasks. Saliency asymmetry was consistently observed in different model architectures and encoder backbones, undermining the reliability of channel-wise interpretability.
A mitigation strategy was proposed --- uniform initialization of all channels using the pretrained green channel --- which effectively reduced asymmetry while maintaining the advantages of pretrained weights. The resulting saliency distributions were more balanced, improving the interpretability of the model predictions.
These findings highlight the importance of carefully adapting pretrained models for specialized datasets, particularly in 2D+ electron microscopy imaging, to ensure robust segmentation performance and reliable attention-based explanations.

{\small
\bibliographystyle{ieee_fullname}

\bibliography{main}

@String(ECCV= {Eur. Conf. Comput. Vis.})

@String(ICPR = {Int. Conf. Pattern Recog.})

@String(ICIP = {IEEE Int. Conf. Image Process.})

@String(ECCV  = {ECCV})

@String(ICPR  = {ICPR})

@String(ICIP  = {ICIP})

@INPROCEEDINGS{ImageNet,
  author={Deng, Jia and Dong, Wei and Socher, Richard and Li, Li-Jia and Kai Li and Li Fei-Fei},
  booktitle={2009 IEEE Conference on Computer Vision and Pattern Recognition}, 
  title={ImageNet: A large-scale hierarchical image database}, 
  year={2009},
  volume={},
  number={},
  pages={248-255},
  doi={10.1109/CVPR.2009.5206848}}

@article{arganda2013snemi3d,
  title={SNEMI3D: 3D Segmentation of neurites in EM images},
  author={Arganda-Carreras, Ignacio and Seung, H Sebastian and Vishwanathan, Ashwin and Berger, Daniel R},
  journal={(No Title)},
  year={2013},
  publisher={Zenodo}
}

@article{lucchi2011supervoxel,
  title={Supervoxel-based segmentation of mitochondria in em image stacks with learned shape features},
  author={Lucchi, Aur{\'e}lien and Smith, Kevin and Achanta, Radhakrishna and Knott, Graham and Fua, Pascal},
  journal={IEEE transactions on medical imaging},
  volume={31},
  number={2},
  pages={474--486},
  year={2011},
  publisher={IEEE}
}

@ARTICLE{pa66,
    AUTHOR={Bertoldo, João P. C. and Decencière, Etienne and Ryckelynck, David and Proudhon, Henry},
    TITLE={A Modular U-Net for Automated Segmentation of X-Ray Tomography Images in Composite Materials},
    JOURNAL={Frontiers in Materials},
    VOLUME={8},
    YEAR={2021},
    URL={https://www.frontiersin.org/article/10.3389/fmats.2021.761229},
    DOI={10.3389/fmats.2021.761229},
    ISSN={2296-8016},
}

@book{fairchild2013color,
  title={Color appearance models},
  author={Fairchild, Mark D},
  year={2013},
  publisher={John Wiley \& Sons}
}

@article{othman2022automatic,
  title={Automatic detection of liver cancer using hybrid pre-trained models},
  author={Othman, Esam and Mahmoud, Muhammad and Dhahri, Habib and Abdulkader, Hatem and Mahmood, Awais and Ibrahim, Mina},
  journal={Sensors},
  volume={22},
  number={14},
  pages={5429},
  year={2022},
  publisher={MDPI}
}

@article{messaoudi2023cross,
  title={Cross-dimensional transfer learning in medical image segmentation with deep learning},
  author={Messaoudi, Hicham and Belaid, Ahror and Salem, Douraied Ben and Conze, Pierre-Henri},
  journal={Medical image analysis},
  volume={88},
  pages={102868},
  year={2023},
  publisher={Elsevier}
}

@inproceedings{roth2014new,
  title={A new 2.5 D representation for lymph node detection using random sets of deep convolutional neural network observations},
  author={Roth, Holger R and Lu, Le and Seff, Ari and Cherry, Kevin M and Hoffman, Joanne and Wang, Shijun and Liu, Jiamin and Turkbey, Evrim and Summers, Ronald M},
  booktitle={Medical Image Computing and Computer-Assisted Intervention--MICCAI 2014: 17th International Conference, Boston, MA, USA, September 14-18, 2014, Proceedings, Part I 17},
  pages={520--527},
  year={2014},
  organization={Springer}
}

@inproceedings{xia2018bridging,
  title={Bridging the gap between 2d and 3d organ segmentation with volumetric fusion net},
  author={Xia, Yingda and Xie, Lingxi and Liu, Fengze and Zhu, Zhuotun and Fishman, Elliot K and Yuille, Alan L},
  booktitle={Medical Image Computing and Computer Assisted Intervention--MICCAI 2018: 21st International Conference, Granada, Spain, September 16-20, 2018, Proceedings, Part IV 11},
  pages={445--453},
  year={2018},
  organization={Springer}
}

@inproceedings{xing20192,
  title={2.5 D convolution for RGB-D semantic segmentation},
  author={Xing, Yajie and Wang, Jingbo and Chen, Xiaokang and Zeng, Gang},
  booktitle={2019 IEEE International Conference on Image Processing (ICIP)},
  pages={1410--1414},
  year={2019},
  organization={IEEE}
}

@article{avesta2023comparing,
  title={Comparing 3D, 2.5 D, and 2D approaches to brain image auto-segmentation},
  author={Avesta, Arman and Hossain, Sajid and Lin, MingDe and Aboian, Mariam and Krumholz, Harlan M and Aneja, Sanjay},
  journal={Bioengineering},
  volume={10},
  number={2},
  pages={181},
  year={2023},
  publisher={MDPI}
}

@article{zhang2022bridging,
  title={Bridging 2D and 3D segmentation networks for computation-efficient volumetric medical image segmentation: An empirical study of 2.5 D solutions},
  author={Zhang, Yichi and Liao, Qingcheng and Ding, Le and Zhang, Jicong},
  journal={Computerized Medical Imaging and Graphics},
  volume={99},
  pages={102088},
  year={2022},
  publisher={Elsevier}
}

@article{yosinski2014transferable,
  title={How transferable are features in deep neural networks?},
  author={Yosinski, Jason and Clune, Jeff and Bengio, Yoshua and Lipson, Hod},
  journal={Advances in neural information processing systems},
  volume={27},
  year={2014}
}

@inproceedings{kornblith2019better,
  title={Do better imagenet models transfer better?},
  author={Kornblith, Simon and Shlens, Jonathon and Le, Quoc V},
  booktitle={Proceedings of the IEEE/CVF conference on computer vision and pattern recognition},
  pages={2661--2671},
  year={2019}
}

@inproceedings{renggli2022model,
  title={Which model to transfer? finding the needle in the growing haystack},
  author={Renggli, Cedric and Pinto, Andr{\'e} Susano and Rimanic, Luka and Puigcerver, Joan and Riquelme, Carlos and Zhang, Ce and Lu{\v{c}}i{\'c}, Mario},
  booktitle={Proceedings of the IEEE/CVF Conference on Computer Vision and Pattern Recognition},
  pages={9205--9214},
  year={2022}
}

@article{zhuang2020comprehensive,
  title={A comprehensive survey on transfer learning},
  author={Zhuang, Fuzhen and Qi, Zhiyuan and Duan, Keyu and Xi, Dongbo and Zhu, Yongchun and Zhu, Hengshu and Xiong, Hui and He, Qing},
  journal={Proceedings of the IEEE},
  volume={109},
  number={1},
  pages={43--76},
  year={2020},
  publisher={Ieee}
}

@article{wald1964receptors,
  title={The Receptors of Human Color Vision: Action spectra of three visual pigments in human cones account for normal color vision and color-blindness},
  author={Wald, George},
  journal={Science},
  volume={145},
  number={3636},
  pages={1007--1016},
  year={1964},
  publisher={American Association for the Advancement of Science}
}

@inproceedings{deng2009imagenet,
  title={Imagenet: A large-scale hierarchical image database},
  author={Deng, Jia and Dong, Wei and Socher, Richard and Li, Li-Jia and Li, Kai and Fei-Fei, Li},
  booktitle={2009 IEEE conference on computer vision and pattern recognition},
  pages={248--255},
  year={2009},
  organization={Ieee}
}

@inproceedings{zeiler2014visualizing,
  title={Visualizing and understanding convolutional networks},
  author={Zeiler, Matthew D and Fergus, Rob},
  booktitle={Computer Vision--ECCV 2014: 13th European Conference, Zurich, Switzerland, September 6-12, 2014, Proceedings, Part I 13},
  pages={818--833},
  year={2014},
  organization={Springer}
}

@inproceedings{selvaraju2017grad,
  title={Grad-cam: Visual explanations from deep networks via gradient-based localization},
  author={Selvaraju, Ramprasaath R and Cogswell, Michael and Das, Abhishek and Vedantam, Ramakrishna and Parikh, Devi and Batra, Dhruv},
  booktitle={Proceedings of the IEEE international conference on computer vision},
  pages={618--626},
  year={2017}
}

@inproceedings{chattopadhay2018grad,
  title={Grad-cam++: Generalized gradient-based visual explanations for deep convolutional networks},
  author={Chattopadhay, Aditya and Sarkar, Anirban and Howlader, Prantik and Balasubramanian, Vineeth N},
  booktitle={2018 IEEE winter conference on applications of computer vision (WACV)},
  pages={839--847},
  year={2018},
  organization={IEEE}
}

@inproceedings{mokuwe2020black,
  title={Black-box saliency map generation using bayesian optimisation},
  author={Mokuwe, Mamuku and Burke, Michael and Bosman, Anna Sergeevna},
  booktitle={2020 International Joint Conference on Neural Networks (IJCNN)},
  pages={1--8},
  year={2020},
  organization={IEEE}
}

@inproceedings{li2019dense,
  title={Dense transformer networks for brain electron microscopy image segmentation},
  author={Li, Jun and Chen, Yongjun and Cai, Lei and Davidson, Ian and Ji, Shuiwang},
  booktitle={Proceedings of the 28th International Joint Conference on Artificial Intelligence},
  year={2019}
}

@inproceedings{matejek2019biologically,
  title={Biologically-constrained graphs for global connectomics reconstruction},
  author={Matejek, Brian and Haehn, Daniel and Zhu, Haidong and Wei, Donglai and Parag, Toufiq and Pfister, Hanspeter},
  booktitle={Proceedings of the IEEE/CVF conference on computer vision and pattern recognition},
  pages={2089--2098},
  year={2019}
}

@inproceedings{gonda2021consistent,
  title={Consistent recurrent neural networks for 3d neuron segmentation},
  author={Gonda, Felix and Wei, Donglai and Pfister, Hanspeter},
  booktitle={2021 IEEE 18th International Symposium on Biomedical Imaging (ISBI)},
  pages={1012--1016},
  year={2021},
  organization={IEEE}
}

@article{januszewski2019segmentation,
  title={Segmentation-enhanced cyclegan},
  author={Januszewski, Micha{\l} and Jain, Viren},
  journal={bioRxiv},
  pages={548081},
  year={2019},
  publisher={Cold Spring Harbor Laboratory}
}

@article{kullback1951information,
  title={On information and sufficiency},
  author={Kullback, Solomon and Leibler, Richard A.},
  journal={The Annals of Mathematical Statistics},
  volume={22},
  number={1},
  pages={79--86},
  year={1951},
  publisher={Institute of Mathematical Statistics}
}

@article{lin1991divergence,
  title={Divergence measures based on the Shannon entropy},
  author={Lin, Jianhua},
  journal={IEEE Transactions on Information theory},
  volume={37},
  number={1},
  pages={145--151},
  year={1991},
  publisher={IEEE}
}

@article{bhattacharyya1943measure,
  title={On a measure of divergence between two statistical populations defined by their probability distributions},
  author={Bhattacharyya, Anil Kumar},
  journal={Bulletin of the Calcutta Mathematical Society},
  volume={35},
  pages={99--109},
  year={1943}
}

@article{bhattacharyya1946measure,
  title={On a measure of divergence between two multinomial populations},
  author={Bhattacharyya, Anil Kumar},
  journal={Sankhyā: The Indian Journal of Statistics},
  volume={7},
  number={4},
  pages={401--406},
  year={1946}
}

@article{kantorovich1942translocation,
  title={On the translocation of masses},
  author={Kantorovich, Leonid V.},
  journal={Doklady Akademii Nauk SSSR},
  volume={37},
  number={7},
  pages={199--201},
  year={1942}
}

@inproceedings{chen2018encoder,
  title={Encoder-decoder with atrous separable convolution for semantic image segmentation},
  author={Chen, Liang-Chieh and Zhu, Yukun and Papandreou, George and Schroff, Florian and Adam, Hartwig},
  booktitle={Proceedings of the European conference on computer vision (ECCV)},
  pages={801--818},
  year={2018}
}

@misc{kingma2017adammethodstochasticoptimization,
      title={Adam: A Method for Stochastic Optimization}, 
      author={Diederik P. Kingma and Jimmy Ba},
      year={2017},
      eprint={1412.6980},
      archivePrefix={arXiv},
      primaryClass={cs.LG},
      url={https://arxiv.org/abs/1412.6980}, 
}

@inproceedings{he2016deep,
  title={Deep residual learning for image recognition},
  author={He, Kaiming and Zhang, Xiangyu and Ren, Shaoqing and Sun, Jian},
  booktitle={Proceedings of the IEEE conference on computer vision and pattern recognition},
  pages={770--778},
  year={2016}
}

@inproceedings{ronneberger2015u,
  title={U-net: Convolutional networks for biomedical image segmentation},
  author={Ronneberger, Olaf and Fischer, Philipp and Brox, Thomas},
  booktitle={Medical image computing and computer-assisted intervention--MICCAI 2015: 18th international conference, Munich, Germany, October 5-9, 2015, proceedings, part III 18},
  pages={234--241},
  year={2015},
  organization={Springer}
}

@article{xie2021segformer,
  title={SegFormer: Simple and efficient design for semantic segmentation with transformers},
  author={Xie, Enze and Wang, Wenhai and Yu, Zhiding and Anandkumar, Anima and Alvarez, Jose M and Luo, Ping},
  journal={Advances in neural information processing systems},
  volume={34},
  pages={12077--12090},
  year={2021}
}

@article{vu2020evaluation,
  title={Evaluation of multislice inputs to convolutional neural networks for medical image segmentation},
  author={Vu, Minh H and Grimbergen, Guus and Nyholm, Tufve and L{\"o}fstedt, Tommy},
  journal={Medical Physics},
  volume={47},
  number={12},
  pages={6216--6231},
  year={2020},
  publisher={Wiley Online Library}
}

@article{amann2020explainability,
  title={Explainability for artificial intelligence in healthcare: a multidisciplinary perspective},
  author={Amann, Julia and Blasimme, Alessandro and Vayena, Effy and Frey, Dietmar and Madai, Vince I and Precise4Q Consortium},
  journal={BMC medical informatics and decision making},
  volume={20},
  pages={1--9},
  year={2020},
  publisher={Springer}
}

@article{dwivedi2023explainable,
  title={Explainable AI (XAI): Core ideas, techniques, and solutions},
  author={Dwivedi, Rudresh and Dave, Devam and Naik, Het and Singhal, Smiti and Omer, Rana and Patel, Pankesh and Qian, Bin and Wen, Zhenyu and Shah, Tejal and Morgan, Graham and others},
  journal={ACM Computing Surveys},
  volume={55},
  number={9},
  pages={1--33},
  year={2023},
  publisher={ACM New York, NY}
}

@article{lee2024foundation,
  title={Foundation models for biomedical image segmentation: A survey},
  author={Lee, Ho Hin and Gu, Yu and Zhao, Theodore and Xu, Yanbo and Yang, Jianwei and Usuyama, Naoto and Wong, Cliff and Wei, Mu and Landman, Bennett A and Huo, Yuankai and others},
  journal={arXiv preprint arXiv:2401.07654},
  year={2024}
}

@article{bahador2025vision,
  title={Vision Transformers Exhibit Human-Like Biases: Evidence of Orientation and Color Selectivity, Categorical Perception, and Phase Transitions},
  author={Bahador, Nooshin},
  journal={arXiv preprint arXiv:2504.09393},
  year={2025}
}

@article{de2022emergent,
  title={Emergent color categorization in a neural network trained for object recognition},
  author={de Vries, Jelmer P and Akbarinia, Arash and Flachot, Alban and Gegenfurtner, Karl R},
  journal={Elife},
  volume={11},
  pages={e76472},
  year={2022},
  publisher={eLife Sciences Publications Limited}
}

@article{rafegas2018color,
  title={Color encoding in biologically-inspired convolutional neural networks},
  author={Rafegas, Ivet and Vanrell, Maria},
  journal={Vision research},
  volume={151},
  pages={7--17},
  year={2018},
  publisher={Elsevier}
}

@article{peddie2022volume,
  title={Volume electron microscopy},
  author={Peddie, Christopher J and Genoud, Christel and Kreshuk, Anna and Meechan, Kimberly and Micheva, Kristina D and Narayan, Kedar and Pape, Constantin and Parton, Robert G and Schieber, Nicole L and Schwab, Yannick and others},
  journal={Nature Reviews Methods Primers},
  volume={2},
  number={1},
  pages={51},
  year={2022},
  publisher={Nature Publishing Group UK London}
}

@article{de2021deciphering,
  title={Deciphering tumour tissue organization by 3D electron microscopy and machine learning},
  author={de Senneville, Baudouin Denis and Khoubai, Fatma Zohra and Bevilacqua, Marc and Labedade, Alexandre and Flosseau, Kathleen and Chardot, Christophe and Branchereau, Sophie and Ripoche, Jean and Cairo, Stefano and Gontier, Etienne and others},
  journal={Communications biology},
  volume={4},
  number={1},
  pages={1390},
  year={2021},
  publisher={Nature Publishing Group UK London}
}

@article{hanslovsky2017image,
  title={Image-based correction of continuous and discontinuous non-planar axial distortion in serial section microscopy},
  author={Hanslovsky, Philipp and Bogovic, John A and Saalfeld, Stephan},
  journal={Bioinformatics},
  volume={33},
  number={9},
  pages={1379--1386},
  year={2017},
  publisher={Oxford University Press}
}

@inproceedings{szabo2022mitigating,
  title={Mitigating the bias of centered objects in common datasets},
  author={Szab{\'o}, Gergely and Horv{\'a}th, Andr{\'a}s},
  booktitle={2022 26th International Conference on Pattern Recognition (ICPR)},
  pages={4786--4792},
  year={2022},
  organization={IEEE}
}

@inproceedings{kayhan2020translation,
  title={On translation invariance in cnns: Convolutional layers can exploit absolute spatial location},
  author={Kayhan, Osman Semih and Gemert, Jan C van},
  booktitle={Proceedings of the IEEE/CVF conference on computer vision and pattern recognition},
  pages={14274--14285},
  year={2020}
}
}

\end{document}